\documentclass[10pt]{article}

\usepackage{times}
\usepackage{epsfig}
\usepackage{graphicx}
\usepackage{amsmath,amssymb}

\usepackage{algorithm}
\usepackage{listings}
\usepackage[table]{xcolor}
\usepackage{booktabs} %
\usepackage{colortbl}
\usepackage{glossaries}
\usepackage{multirow}
\usepackage[font=small]{caption}
\usepackage{palatino}

\usepackage{pifont}% http://ctan.org/pkg/pifont
\usepackage{xspace}

\usepackage[pagebackref=true,breaklinks=true,letterpaper=true,colorlinks,bookmarks=false]{hyperref}

\newif\ifarxiv

\definecolor{myblue}{RGB}{8,48,107}
\definecolor{myred}{RGB}{127,39,4}

\newcommand{\mycolrule}{\arrayrulecolor{black!30} \midrule \arrayrulecolor{black}}

\newcommand \myfig[1] {\includegraphics[width=0.325\linewidth,height=0.325\linewidth]{figs/images/#1}}
\newcommand \myfigc[1] {\includegraphics[trim=7 6 7 7,clip,width=0.325\linewidth,height=0.325\linewidth,]{figs/images/#1}}

\def \pzo {\phantom{0}} 
\def \dzo {\phantom{00}}

\def \SA {\mathrm{SA}\xspace}
\def \FFN {\mathrm{FFN}\xspace}
\def \LN {\eta\xspace}
\def \alambic {$\Upsilon$\xspace}
\def \alambicb {$\Upsilon$\xspace}

\def \OURS {CaiT\xspace}

\def \etal {\textit{et al.}\xspace}

\def \bestperf {86.5\%\xspace}

\def \ours {\OURS}

\author{
\begin{minipage}{\linewidth}
\begin{center}
\large Hugo Touvron$^{\star,\dagger}$ \hspace{0.25cm} Matthieu Cord$^{\dagger}$ \hspace{0.25cm} Alexandre Sablayrolles$^{\star}$ \\[0.2cm] 
Gabriel Synnaeve$^{\star}$ \hspace{0.3cm} Herv\'e J\'egou$^{\star}$ \\[0.5cm]
\scalebox{1.}{$^\star$Facebook AI\hspace{0.6cm} $^\dagger$Sorbonne University}\\[1cm]
\end{center}
\end{minipage}
}
\title{Going deeper with Image Transformers } 

\makeatletter
\let\inserttitle\@title
\makeatother

\makeatletter
\renewcommand{\paragraph}{%
  \@startsection{paragraph}{4}%
  {\z@}{2.25ex \@plus 1ex \@minus .2ex}{-1em}%
  {\normalfont\normalsize\bfseries}%
}
\makeatother

\date{~}
\begin{document}

\maketitle

\begin{abstract}
    Transformers have been recently adapted for large scale image classification, achieving high scores shaking up the long supremacy of convolutional neural networks.
    However the optimization of image transformers has been little studied so far.
    In this work, we build and optimize deeper transformer networks for image classification.
    In particular, we investigate the interplay of architecture and optimization of such dedicated transformers. 
    We make two transformers architecture changes that significantly improve the accuracy of deep transformers. This leads us to produce models whose performance does not saturate early with more depth, for instance we obtain \bestperf top-1 accuracy on Imagenet when training with no external data, we thus attain the current SOTA with less FLOPs and parameters. Moreover, our best model establishes the new state of the art on Imagenet with Reassessed labels and Imagenet-V2 / match frequency, in the setting with no additional training data. We share our code and models\footnote{\url{https://github.com/facebookresearch/deit}}. 
\end{abstract}

\section{Introduction}
\label{sec:introduction}

Residual architectures are prominent in computer vision since the advent of ResNet~\cite{He2016ResNet}. They are defined as a sequence of functions of the form 
\begin{equation}
    x_{l+1} = g_l(x_l) + R_l(x_l),
\end{equation}
where the function $g_l$ and $R_l$ define how the network updates the input $x_l$ at layer $l$. The function $g_l$ is typically identity, while $R_l$ is the main building block of the network: 
many variants in the literature essentially differ on how one defines this residual branch $R_l$ is constructed or parametrized~\cite{Radosavovic2020RegNet,tan2019efficientnet,Xie2017AggregatedRT}. 
Residual architectures highlight the strong interplay between optimization and architecture design. As pointed out by He \etal~\cite{He2016ResNet}, residual networks do not offer a better representational power. 
They achieve better performance because they are easier to train: 
shortly after their seminal work, He \etal~discussed~\cite{He2016IdentityMappings} the importance of having a clear path both forward and backward, and advocate 
setting $g_l$ to the identity function. %

The vision transformers~\cite{dosovitskiy2020image} instantiate a particular form of residual architecture: after casting the input image into a set $x_0$ of vectors, the network alternates self-attention layers ($\SA$) with feed-forward networks ($\FFN$), as 
\begin{align}
    x'_{l}  & = x_{l} + \SA(\LN(x_l))  \nonumber \\
    x_{l+1} & = x'_{l} + \FFN(\LN(x'_l))
    \label{equ:transformer}
\end{align}
where $\eta$ is the LayerNorm operator~\cite{ba2016layer}. 
This definition follows the original architecture of Vaswani \etal~\cite{vaswani2017attention}, except the LayerNorm is applied before the block (\emph{pre-norm}) in the residual branch, as advocated by He \etal\cite{He2016IdentityMappings}. Child~\etal~\cite{child2019generating} adopt this choice with LayerNorm for training deeper transformers for various media, including for image generation where they train transformers with 48 layers.  

How to normalize, weigh, or initialize the residual blocks of a residual architecture has received  significant attention both for convolutional neural networks~\cite{brock2021characterizing,Brock2021HighPerformanceLI,He2016IdentityMappings,Zhang2019FixupIR} and for transformers applied to NLP or speech tasks~\cite{Bachlechner2020ReZeroIA,huang2020improving,Zhang2019FixupIR}. 
In Section~\ref{sec:training}, we revisit this topic for transformer architectures solving image classification problems. 
Examples of approaches closely related to ours include  Fixup~\cite{Zhang2019FixupIR}, T-Fixup~\cite{huang2020improving},  ReZero~\cite{Bachlechner2020ReZeroIA} and   SkipInit~\cite{de2020batch}. 

Following our analysis of the interplay between different initialization, optimization and architectural design, we propose an approach that is effective to improve the training of deeper architecture compared to current methods for image transformers. Formally, we add a learnable diagonal matrix on output of each residual block, initialized close to (but not at) 0. 
Adding this simple layer after each residual block improves the training dynamic, allowing us to train deeper high-capacity image transformers that benefit from depth. 
We refer to this approach as \textbf{LayerScale}. 

Section~\ref{sec:method} introduces our second contribution, namely \textbf{class-attention layers,} that we present in Figure~\ref{fig:mix_and_summarize}. It is akin to an encoder/decoder architecture, in which we explicitly separate the transformer layers involving self-attention between patches, from class-attention layers that are devoted to extract the content of the processed patches into a single vector so that it can be fed to a linear classifier.  
This explicit separation avoids the contradictory objective of guiding the attention process while processing the class embedding. We refer to this new architecture as \textbf{\OURS} (Class-Attention in Image Transformers). 

In the experimental Section~\ref{sec:experiments}, we empirically show the effectiveness and complementary of our approaches: 
\begin{itemize}
    \item LayerScale significantly facilitates the convergence and improves the accuracy of image transformers at larger depths. It adds a few thousands of parameters to the network at training time (negligible w.r.t. the total number of weights). % 
    
    \item Our architecture with specific class-attention offers a more effective processing of the class embedding. 

    \item Our best \OURS models establish the new state of the art on Imagenet-Real~\cite{Beyer2020ImageNetReal} and Imagenet V2 matched  frequency~\cite{Recht2019ImageNetv2} with no additional training data. On ImageNet1k-val~\cite{Russakovsky2015ImageNet12}, our model is on par with the state of the art (\bestperf) while requiring less FLOPs (329B vs 377B) and having less parameters than the best competing model (356M vs 438M). %. 

    \item We achieve competitive results on Transfer Learning. %
\end{itemize}

We provide visualizations of the attention mechanisms in Section~\ref{sec:vizualizations}. We discuss related works along this paper and in the dedicated Section~\ref{sec:related}, before we conclude in Section~\ref{sec:conclusion}. The appendices contain some variations we have tried during our exploration. 

\section{Deeper image transformers with LayerScale}
\label{sec:training}

Our goal is to increase the stability of the optimization when training transformers for image classification derived from the original architecture by Vaswani \etal.~\cite{vaswani2017attention}, and especially when we increase their depth. 
We consider more specifically the vision transformer (ViT) architecture proposed by Dosovitskiy \etal~\cite{dosovitskiy2020image} as the reference architecture and adopt the data-efficient image transformer (DeiT) optimization procedure of Touvron \etal~\cite{Touvron2020TrainingDI}. 
In both works, there is no evidence that depth can bring any benefit when training on Imagenet only: the deeper ViT architectures have a low performance, while DeiT only considers transformers with 12 blocks of layers. 
The experimental section~\ref{sec:experiments} will confirm that DeiT does not train deeper models effectively.  

\begin{figure}
    %\centering
    \includegraphics[width=1\linewidth]{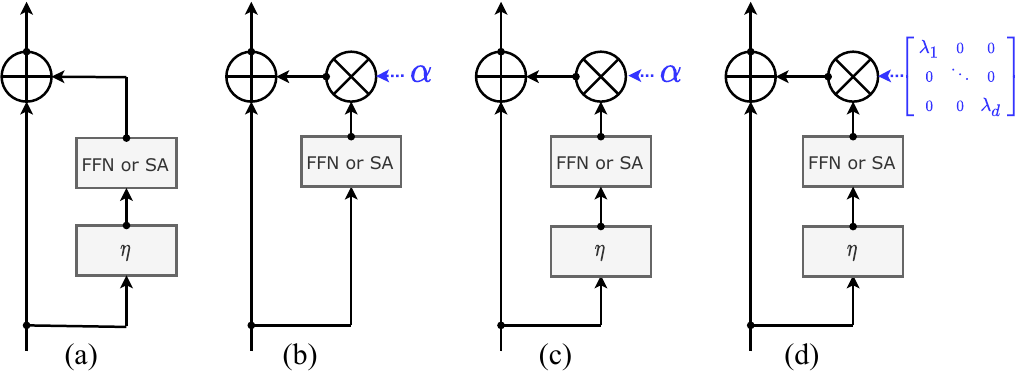}  
    \caption{Normalization strategies for transformer blocks. (a) The ViT image classifier adopts pre-normalization like Child \etal~\cite{child2019generating}. (b)  ReZero/Skipinit and Fixup remove the $\eta$ normalization and the warmup (i.e., a reduced learning rate in the early training stage) and add a learnable scalar initialized to $\alpha$\,=\,$0$ and $\alpha$\,=\,$1$, respectively. Fixup additionally introduces biases and modifies the initialization of the linear layers. Since these methods do not converge with deep vision transformers, (c) we adapt them by re-introducing the pre-norm $\eta$ and the warmup. Our main proposal (d) introduces a per-channel weighting (i.e, multiplication with a diagonal matrix $\mathrm{diag}(\lambda_1,\dots,\lambda_d)$, where we initialize each weight with a small value as $\lambda_i = \varepsilon$. 
    \label{fig:normalization}}
\end{figure}

Figure~\ref{fig:normalization} depicts the main variants that we compare for helping the optimization. 
They cover recent choices from the literature:  
as discussed in the introduction, the architecture (a) of ViT and DeiT  is a pre-norm architecture~\cite{dosovitskiy2020image,Touvron2020TrainingDI}, in which the layer-normalisation $\eta$ occurs at the beginning of the residual branch. Note that the original architecture of Vaswani \etal~\cite{vaswani2017attention} applies the normalization after the block, but in our experiments the DeiT training does not converge with post-normalization. 

Fixup~\cite{Zhang2019FixupIR}, ReZero~\cite{Bachlechner2020ReZeroIA} and SkipInit~\cite{de2020batch} introduce learnable scalar weighting $\alpha_l$ on the output of residual blocks, while removing the pre-normalization and the warmup, see Figure~\ref{fig:normalization}(b). This amounts to modifying Eqn.~\ref{equ:transformer} as
\begin{align}
    x'_{l}  & = x_{l} + \alpha_l \, \SA(x_l)  \nonumber \\
    x_{l+1} & = x'_{l} + \alpha'_l \, \FFN(x'_l). 
    \label{equ:transformer3}
\end{align}

ReZero simply initializes this parameter to $\alpha=0$.  
Fixup initializes this parameter to $\alpha=1$ and makes other modifications: it adopts different policies for the initialization of the block weights, and adds several weights to the parametrization. 
In our experiments, these approaches do not converge even with some adjustment of the hyper-parameters.

Our empirical observation is that removing the warmup and the layer-normalization is what makes training unstable in Fixup and T-Fixup.  
Therefore we re-introduce these two ingredients so that Fixup and T-Fixup converge with DeiT models, see Figure~\ref{fig:normalization}(c). 
As we see in the experimental section, these amended variants of Fixup and T-Fixup are effective, mainly due to the learnable parameter $\alpha_l$. When initialized at a small value, this choice does help the convergence when we increase the depth. 

% 

%.

\paragraph{Our proposal LayerScale} is a per-channel multiplication of the vector produced by each residual block, as opposed to a single scalar, see Figure~\ref{fig:normalization}(d). 
Our objective is to 
group the updates of the weights associated with the same output channel. %
Formally, LayerScale is a multiplication by a diagonal matrix on output of each residual block. 
In other terms, we modify Eqn.~\ref{equ:transformer} as 
\begin{align}
    x'_{l}  & = x_{l} + \mathrm{diag}(\lambda_{l,1},\dots,\lambda_{l,d}) \times  \SA(\LN(x_l))  \nonumber \\
    x_{l+1} & = x'_{l} + \mathrm{diag}(\lambda'_{l,1},\dots,\lambda'_{l,d}) \times  \FFN(\LN(x'_l)), 
    \label{equ:transformer2}
\end{align}
where the parameters $\lambda_{l,i}$ and $\lambda'_{l,i}$ are learnable weights. 
The diagonal values are all initialized to a fixed small value $\varepsilon$: we set it to $\varepsilon=0.1$ until depth 18, $\varepsilon=10^{-5}$ for depth 24 and $\varepsilon=10^{-6}$ for deeper networks. 
This formula is akin to other normalization strategies ActNorm~\cite{kingma2018glow} or LayerNorm but executed on output of the residual block. Yet we seek a different effect: ActNorm is a data-dependent initialization that calibrates activations so that they have zero-mean and unit variance, like batchnorm~\cite{ioffe15batchnorm}. In contrast, we initialize the diagonal with small values so that the initial contribution of the residual branches to the function implemented by the transformer is small. In that respect our motivation is therefore closer to that of ReZero~\cite{Bachlechner2020ReZeroIA}, SkipInit~\cite{de2020batch}, Fixup~\cite{Zhang2019FixupIR} and T-Fixup~\cite{huang2020improving}: to train closer to the identity function and let the network integrate the additional parameters progressively during the training. 
LayerScale offers more diversity in the optimization than just adjusting the whole layer by a single learnable scalar as in ReZero/SkipInit, Fixup and T-Fixup. 
As we will show empirically, offering the degrees of freedom to do so per channel is a decisive advantage of LayerScale over existing approaches.In Appendix~\ref{sec:variant_layerscale}, we present other variants or intermediate choices that support our proposal, and a control experiment that aims at disentangling the specific weighting of the branches of LayerScale from its impact on optimization. 

Formally, adding these weights does not change the expressive power of the architecture since, as they can be integrated into the previous matrix of the $\SA$ and $\FFN$ layers without changing the function implemented by the network. 

\section{Specializing layers for class attention}
\label{sec:method}

\begin{figure}[t]
    \centering
    \includegraphics[width=0.9\linewidth]{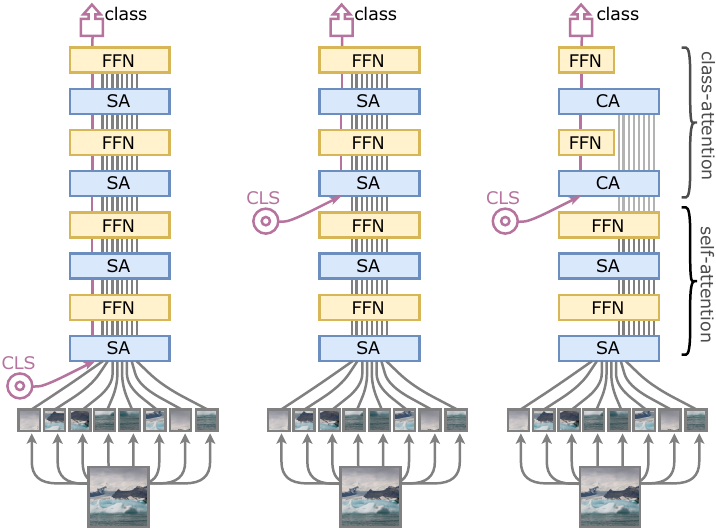}
    \caption{In the ViT transformer (\emph{left}), the class embedding (CLS) is inserted along with the patch embeddings. This choice is detrimental, as the same weights are used for two different purposes: helping the attention process, and preparing the vector to be fed to the classifier. We put this problem in evidence by showing that inserting CLS later improves performance (\emph{middle}). In the \textbf{\OURS} architecture (\emph{right}), we further propose to freeze the patch embeddings when inserting CLS to save compute, so that the last part of the network (typically 2 layers) is fully devoted to summarizing the information to be fed to the linear classifier. 
    }
    \label{fig:mix_and_summarize}
\end{figure}

% \MC{a reprendre en boostant CaiT ici, peut etre revoir les titres S2 et S3}
In this section, we introduce the \OURS architecture, depicted in Figure~\ref{fig:mix_and_summarize} (right). 
This design aims at circumventing one of the problems of the ViT architecture: the learned weights are asked to optimize two contradictory objectives: (1) guiding the self-attention between patches while (2) summarizing the information useful to the linear classifier. Our proposal is to explicitly separate the two stages, in the spirit of an encoder-decoder architecture, see Section~\ref{sec:related}.

\paragraph{Later class token.} 
As an intermediate step towards our proposal, we insert the so-called class token, denoted by CLS, later in the transformer. This choice eliminates the discrepancy on the first layers of the transformer, which are therefore fully employed for performing self-attention between patches only. 
As a baseline that does not suffer from the contradictory objectives, we also consider \textbf{average pooling} of all the patches on output of the transformers, as typically employed in convolutional architectures.

\paragraph{Architecture.} 
Our \OURS network consists of two distinct processing stages visible in Figure~\ref{fig:mix_and_summarize}:
\begin{enumerate} 
\item The \emph{self-attention} stage is identical to the ViT transformer, but with no class embedding (CLS). 
\item The \emph{class-attention} stage is a set of layers that compiles the set of patch embeddings into a class embedding CLS that is subsequently fed to a linear classifier. 
\end{enumerate} 

This class-attention alternates in turn a layer that we refer to as a multi-head class-attention (CA), and a FFN layer.
In this stage, only the class embedding is updated. Similar to the one fed in ViT and DeiT on input of the transformer, it is a learnable vector. The main difference is that, in our architecture, we do no copy information  from the class embedding to the patch embeddings during the forward pass. Only the class embedding is updated by residual in the CA and FFN processing of the class-attention stage. 

\paragraph{Multi-heads class attention.} 
The role of the CA layer is to extract the information from the set of processed patches. %
It is identical to a SA layer, except that it relies on the attention between (i) the class embedding $x_\mathrm{class}$ (initialized at CLS in the first CA) and (ii) itself plus the set of frozen patch embeddings $x_\mathrm{patches}$.  
We discuss why we include $x_\mathrm{class}$ in the keys in Appendix~\ref{sec:variant_class_attention}.

Considering a network with $h$ heads and $p$ patches, and denoting by $d$ the embedding size, we parametrize the multi-head class-attention with several projection matrices,  $W_{q},W_{k},W_{v},W_{o} \in \mathbf{R}^{d \times d}$, and  the corresponding biases  $b_q,b_k,b_v,b_o \in \mathbf{R}^{d}$.
With this notation, the computation of the CA residual block proceeds as follows. We first augment the patch embeddings (in matrix form) as 
$z = [x_\mathrm{class},x_\mathrm{patches}]$   
 (see Appendix~\ref{sec:variant_class_attention} for results when $z=x_\mathrm{patches}$). We then perform the projections:
\begin{align}
Q & = W_{q}  \, x_\mathrm{class} + b_q, \label{equ:ca1} \\
K & = W_{k} \, z + b_k, \label{equ:ca2}\\
V & = W_{v} \, z + b_v. \label{equ:ca3}
\end{align}

The class-attention weights are given by 
\begin{align}
A = \mathrm{Softmax}(Q.K^{T}/ \sqrt{d/h})
\end{align} 
where $Q.K^{T} \in \mathbf{R}^{h \times 1 \times p}$. 
This attention is involved in the weighted sum $A \times V$
to produce the residual output vector
\begin{align}
\mathrm{\mathrm{out}_\mathrm{CA}} =  W_{o} \, A \, V + b_{o},
\end{align}
which is in turn added to $x_\mathrm{class}$ for subsequent processing. 

The CA layers extract the useful information from the patches embedding to the class embedding. In preliminary experiments, we empirically observed that the first CA and FFN give the main boost, and a set of 2 blocks of layers (2 CA and 2 FFN) is sufficient to cap the performance. In the experimental section, we denote by 12+2 a transformer when it consists of 12 blocks of SA+FFN layers and 2 blocks of CA+FFN layers. 

\paragraph{Complexity.} 
The layers contain the same number of parameters in the class-attention and self-attention stages: CA is identical to SA in that respect, and we use the same parametrization for the FFNs. 
However the processing of these layers is much faster: the FFN only processes matrix-vector multiplications. 

The CA function is also less expensive than SA in term of memory and computation because it computes the attention between the class vector and the set of patch embeddings: $Q \in \mathbf{R}^d$ means that $Q.K^{T} \in \mathbf{R}^{h \times 1 \times p}$. In contrast, in the ``regular self-attention'' layers SA, 
we have $Q \in \mathbf{R}^{p \times d}$ and therefore $Q.K^{T} \in \mathbf{R}^{h \times p \times p}$. In other words, the initially quadratic complexity in the number of patches becomes linear  in our extra \OURS layers.

\section{Experiments}
\label{sec:experiments}

In this section, we report our experimental results related to LayerScale and \OURS. 
We first study strategies to train at deeper scale in Section~\ref{sec:analysis_deeper}, including our LayerScale method. Section~\ref{sec:exp_class_attention} shows the interest of our class-attention design.  
We present our models in Subsection~\ref{sec:models}. Section~\ref{sec:results} details our results on Imagenet and Transfer learning. 
We provide an ablation of hyper-parameter and ingredients in Section~\ref{sec:ablation}. 
Note, in Appendix~\ref{sec:variant_layerscale} and~\ref{sec:variant_class_attention}, we provide variations on our methods and corresponding results.  
\paragraph{Experimental setting.} 
Our implementation is based on the Timm library~\cite{pytorchmodels}. Unless specified otherwise, for this analysis we make minimal changes to hyper-parameters compared to the DeiT training scheme~\cite{Touvron2020TrainingDI}.  
In order to speed up training and optimize memory consumption we have used a sharded training provided by the Fairscale library\footnote{\url{https://pypi.org/project/fairscale/}}  
with fp16 precision.

\subsection{Preliminary analysis with deeper architectures} 
\label{sec:analysis_deeper}

In our early experiments, we observe that Vision Transformers become increasingly more difficult to train when we scale architectures. 
Depth is one of the main source of instability. For instance the DeiT procedure~\cite{Touvron2020TrainingDI} fails to properly converge above 18 layers without adjusting hyper-parameters. Large ViT~\cite{dosovitskiy2020image} models with 24 and 32 layers were trained with large training datasets, but when trained on Imagenet only the larger models are not competitive. 

In the following, we analyse various ways to stabilize the training with different architectures. 
 At this stage we consider a Deit-Small model\footnote{\url{https://github.com/facebookresearch/deit}} during 300 epochs to allow a direct comparison with the results reports by Touvron \etal~\cite{Touvron2020TrainingDI}. 
We measure the performance on the  Imagenet1k~\cite{deng2009imagenet,Russakovsky2015ImageNet12} classification dataset as a function of the depth.

\subsubsection{Adjusting the drop-rate of stochastic depth.}  

The first step to improve convergence is to adapt the hyper-parameters that interact the most with depth, in particular Stochastic depth~\cite{Huang2016DeepNW}. This method is already popular in NLP~\cite{fan2019reducing,fan2020training} to train deeper architectures. For ViT, it was first proposed by Wightman \etal~\cite{pytorchmodels} in the Timm implementation, and subsequently adopted in DeiT~\cite{Touvron2020TrainingDI}. The per-layer drop-rate depends linearly on the layer depth, but in our experiments this choice does not provide an advantage compared to the simpler choice of a uniform drop-rate $d_r$.  
In Table~\ref{tab:init_comp} we show that the default stochastic depth of DeiT allows us to train up to 18 blocks of SA+FFN. After that the training becomes unstable. By increasing the drop-rate hyper-parameter  $d_r$, the performance increases until 24 layers. It saturates at 36 layers (we measured that it drops to 80.7\% at 48 layers).

\subsubsection{Comparison of normalization strategies} 
\begin{table}[t]
    \caption{\textbf{Improving convergence at depth} on ImageNet-1k. The baseline is DeiT-S with uniform drop rate of $d=0.05$ (same expected drop rate and performance as progressive stochastic depth of $0.1$). Several methods include a fix scalar learnable weight $\alpha$ per layer as in Figure~\ref{fig:normalization}(c). We have adapted Rezero, Fixup, T-Fixup, since the original methods do not converge: we have re-introduced the Layer-normalization $\eta$ and warmup. We have adapted the drop rate $d_r$ for all the methods, including the baseline. 
    The column $\alpha=\varepsilon$ reports the performance when initializing the scalar with the same value as for LayerScale. $\dagger$: {\it failed before the end of the training. }
    \label{tab:init_comp}}
    %\hspace{-3pt}
    \centering \scalebox{0.9}{
    \begin{tabular}{@{\ }c@{\ }|cc|cccc|c@{\ }}
    \toprule
         \multirow{2}{3em}{\ depth} & \multicolumn{2}{c}{baseline} & 
         \multicolumn{4}{c}{\centering scalar $\alpha$ weighting} & \multirow{2}{*}{LayerScale} \\
           \cmidrule(lr){2-3}  \cmidrule(lr){4-7}
          &  $d_r=0.05$ & adjust [$d_r$] & Rezero  &  T-Fixup  &  Fixup  & $\alpha=\varepsilon$ &  \\         
     %     & adaptation & adaptation & adaptation & weighting & \\
          \midrule
         \pzo12    & 79.9\pzo      & 79.9\  {\small [{\it 0.05}]} & 78.3 & 79.4 & 80.7 & 80.4 & 80.5 \\ 
         \pzo18    & 80.1\pzo      & 80.7\  {\small [{\it 0.10}]} & 80.1 & 81.7 & 82.0 & 81.6 & 81.7 \\ 
         \pzo24    & 78.9$\dagger$ & 81.0\  {\small [{\it 0.20}]} & 80.8 & 81.5 & 82.3 & 81.1 & 82.4 \\ 
         \pzo36    & 78.9$\dagger$ & 81.9\  {\small [{\it 0.25}]} & 81.6 & 82.1 & 82.4 & 81.6 & 82.9 \\ 
          \bottomrule
    \end{tabular}}
\end{table}

We carry out an empirical study of the normalization methods discussed in Section~\ref{sec:training}.
As previously indicated, Rezero, Fixup and T-Fixup do not converge when training DeiT off-the-shelf. However, if we re-introduce LayerNorm\footnote{Bachlechner \etal report that batchnorm is complementary to ReZero, while removing LayerNorm in the case of transformers.} and warmup, Fixup and T-Fixup achieve congervence and even improve training compared to the baseline DeiT. 
We report the results for these ``adaptations'' of Fixup and T-Fixup in Table~\ref{tab:init_comp}.

\begin{figure}[t]
    \centering
    \includegraphics[width=0.6\linewidth]{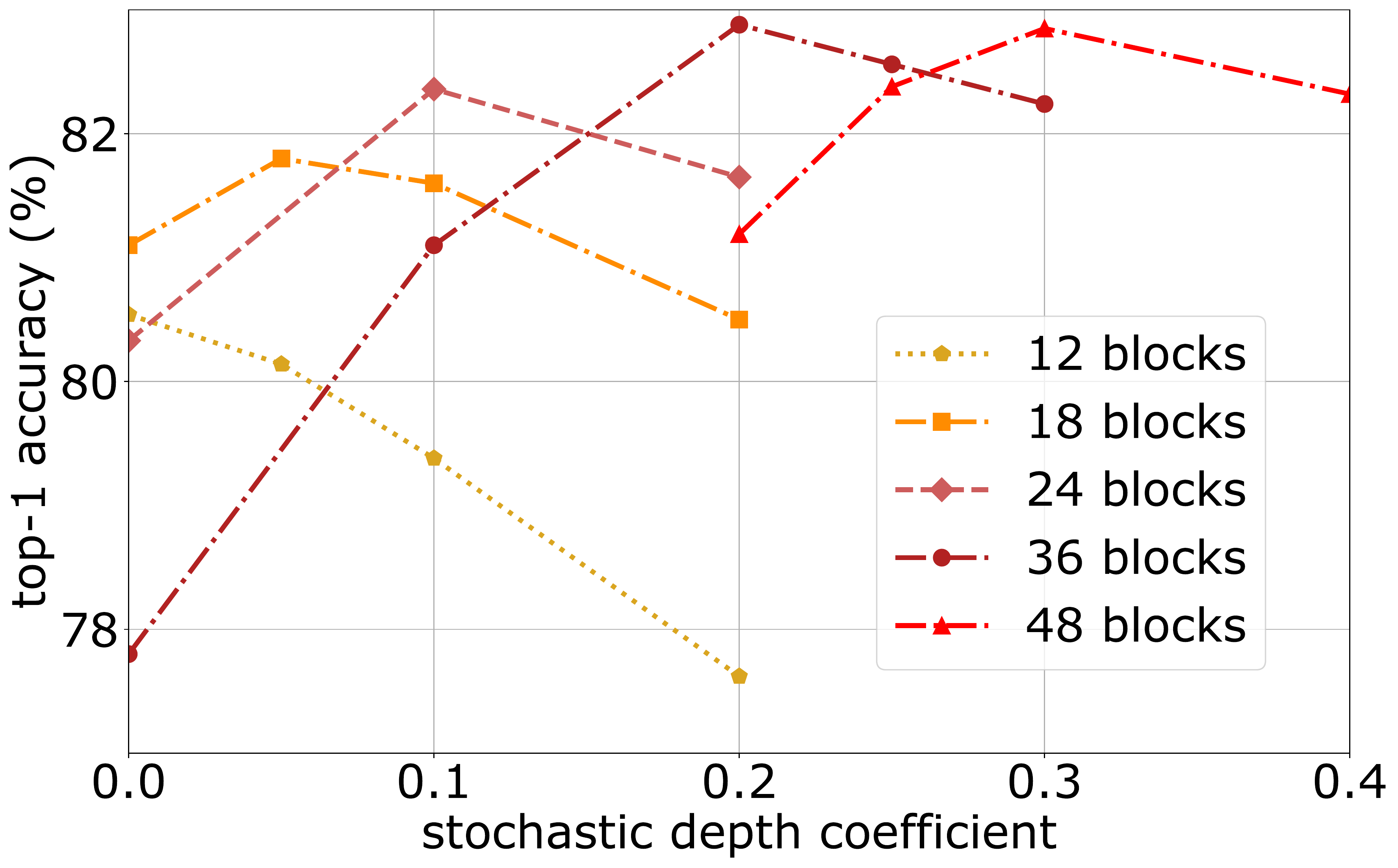}
    \medskip
    \caption{We measure the impact of stochastic depth on ImageNet with a DeiT-S with LayerScale for  different depths. The drop rate of stochastic depth needs to be adapted to the network depth. %\rv{Hugo: do you have 48 layers?} \rv{please use depth for x-axis, and stochastic depth paramters for different lines}
    }
    \label{fig:stch_depth_optim}
\end{figure}

The modified methods are able to converge with more layers without saturating too early. ReZero converges, we show  (column $\alpha=\varepsilon$) that it is better to initialize $\alpha$ to a small value instead of 0, as in LayerScale. 
All the methods have a beneficial effect on convergence and they tend to reduce the need for stochastic depth, therefore we adjust these drop rate accordingly per method. Figure~\ref{fig:stch_depth_optim} provides the performance as the function of the drop rate $d_r$ for LayerScale. We empirically use the following formula to set up the drop-rate for the \OURS-S models derived from on Deit-S:
$%
    d_r$\,=\,$\min(0.1 \times \frac{\mathrm{depth}}{12}-1, 0). 
$%
This formulaic choice avoids cross-validating this parameter and overfitting, yet it does not generalize to models with different $d$: We further increase (resp. decrease) it by a constant for larger (resp. smaller) working dimensionality $d$. 

Fixup and T-Fixup are competitive with LayerScale in the regime of a relatively low number of blocks (12--18). However, they are more complex than LayerScale: they employ different initialization rules depending of the type of layers, and they require more changes to the transformer architecture. Therefore we only use LayerScale in subsequent experiments. It is much simpler and parametrized by a single hyper-parameter $\varepsilon$, and it offers a better performance for the deepest models that we consider, which are also the more accurate.

\subsubsection{Analysis of Layerscale} 

\begin{figure}[t]
\begin{minipage}{0.99\linewidth}
\includegraphics[trim = 0 12 0 0, clip, height=0.37\linewidth]{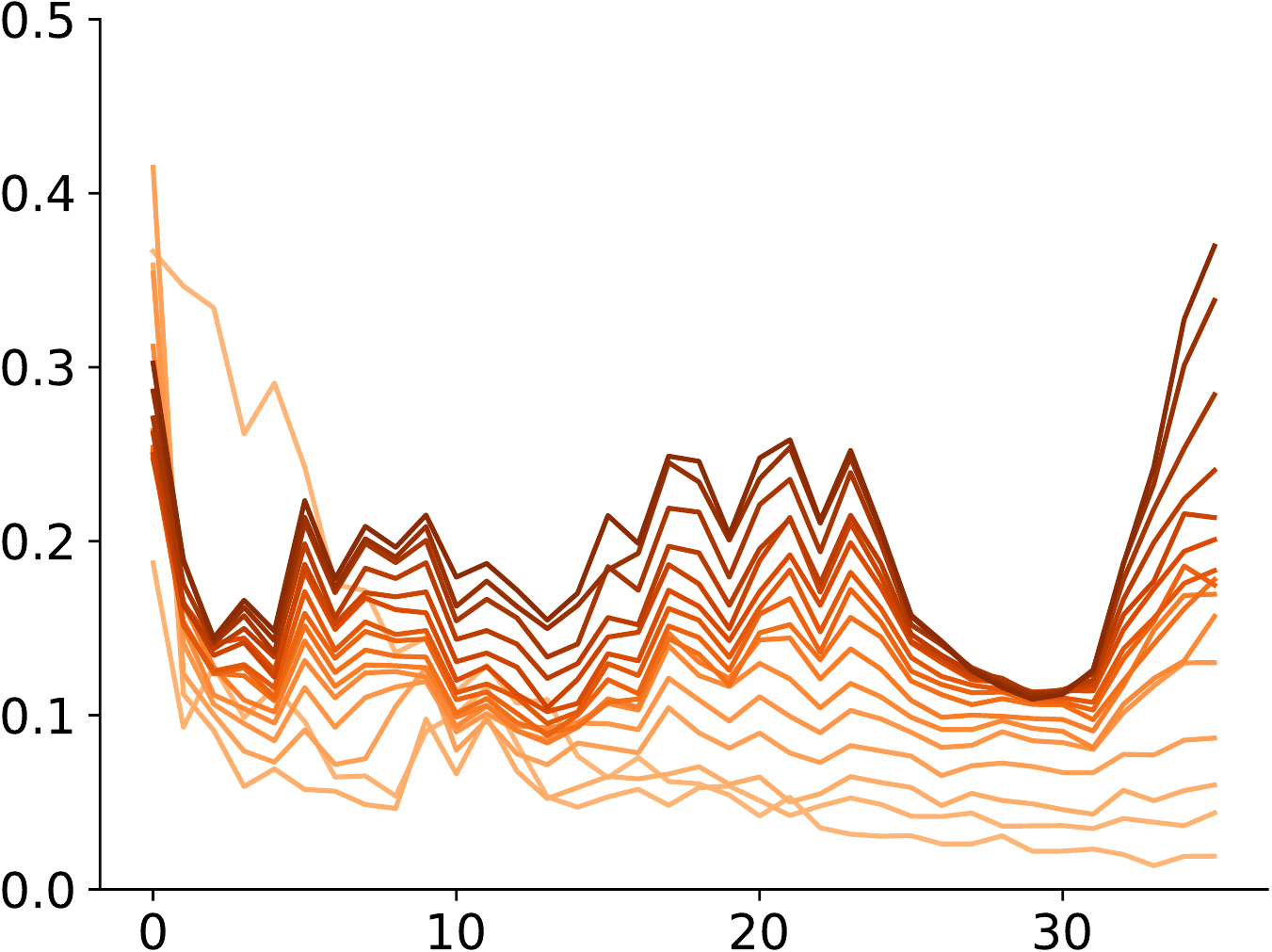}
\hfill
\includegraphics[trim = 40 12 10 0, clip, height=0.37\linewidth]{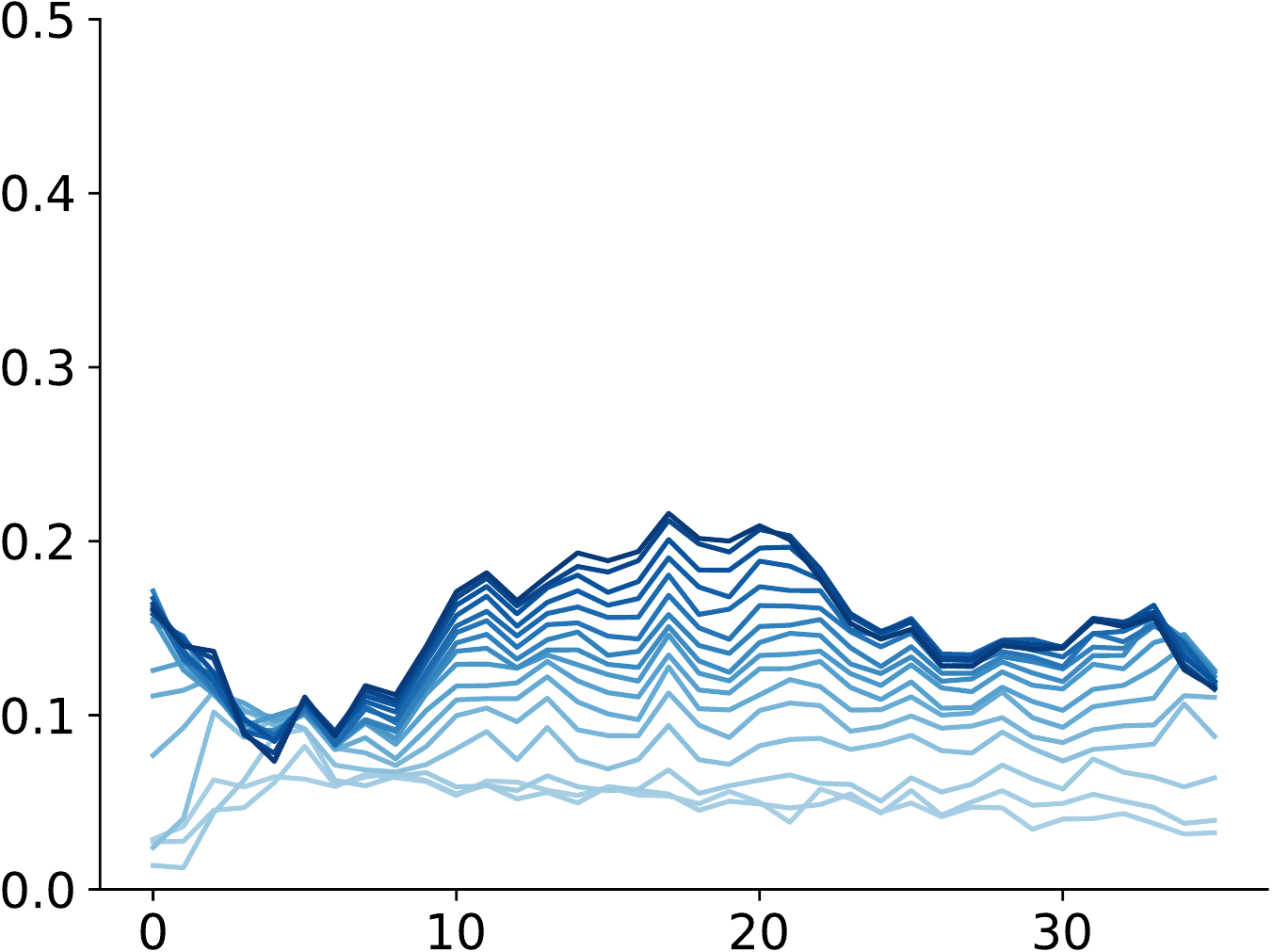}
\end{minipage}%    
\medskip

\begin{minipage}{0.99\linewidth}
\includegraphics[trim = 0 0 0 0, clip, height=0.38\linewidth]{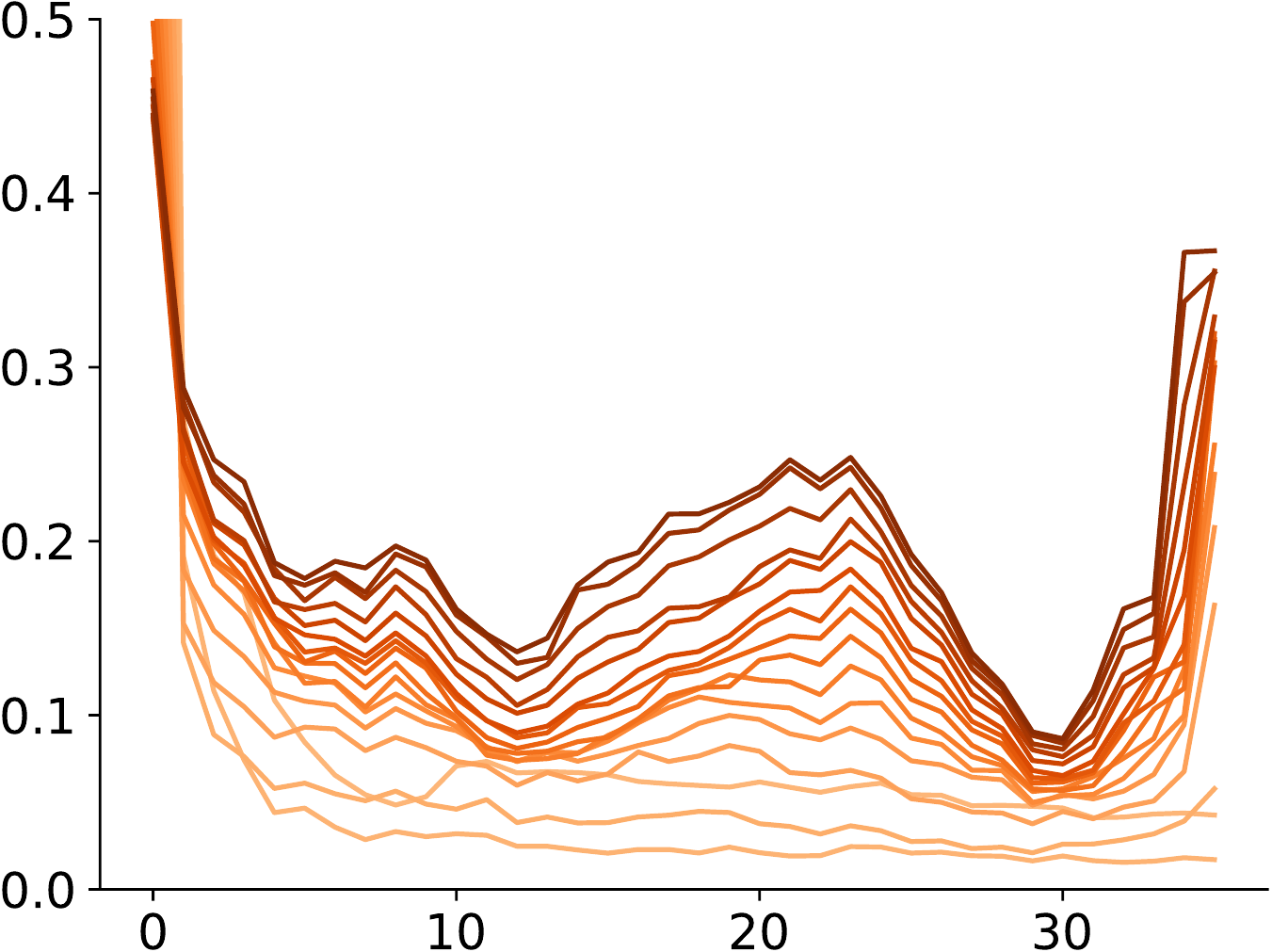}
\hfill
\includegraphics[trim = 40 0 10 0, clip, height=0.38\linewidth]{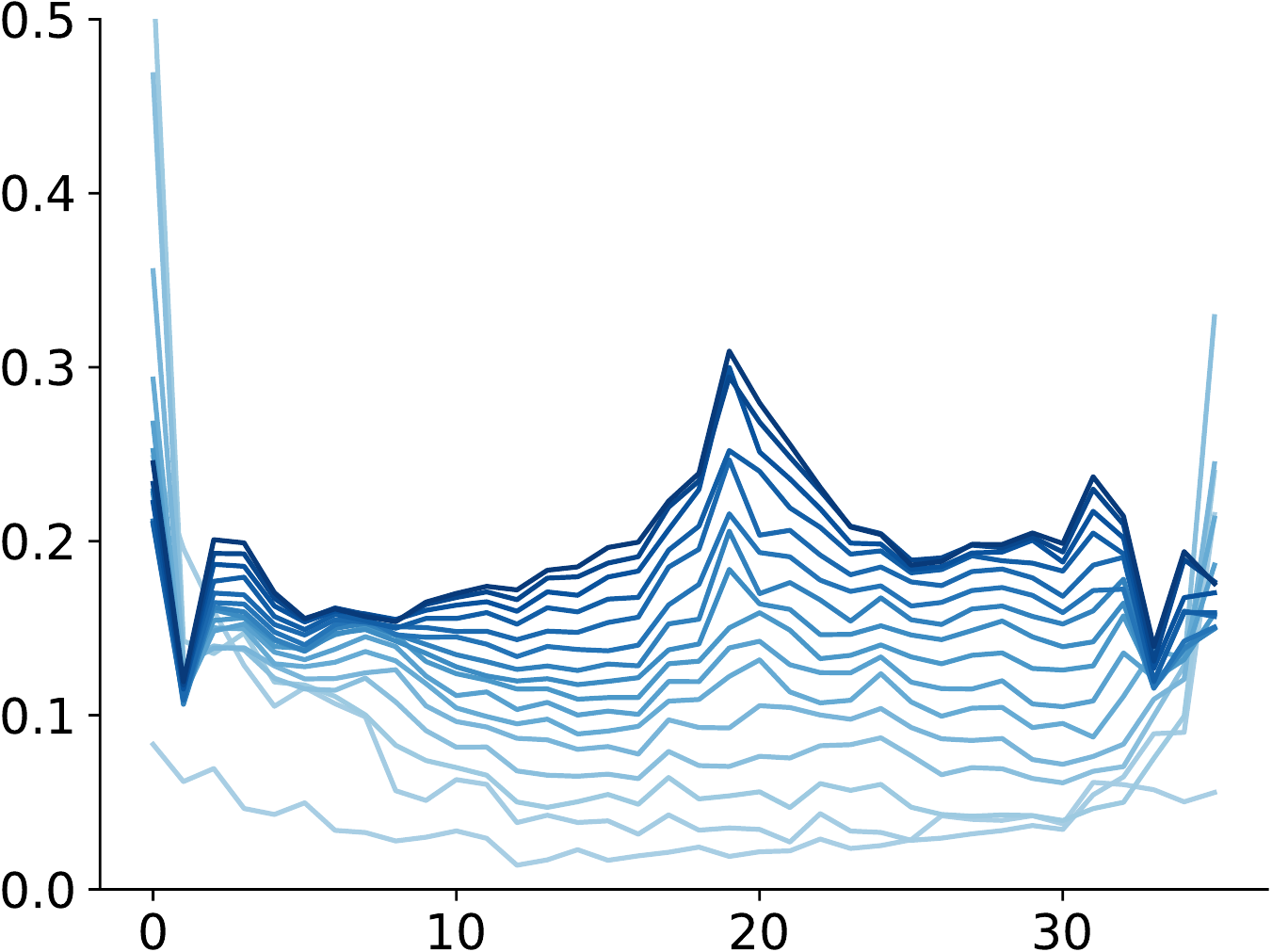}
\end{minipage}%    
\caption{
Analysis of the contribution of the residual branches (\emph{Top:} Self-attention ; \emph{Bottom:} FFN) for a network comprising 36 layers, \textcolor{myred}{without (red)} or \textcolor{myblue}{with (blue)} Layerscale. 
The ratio between the norm of the residual and the norm of the main branch is shown for each layer of the transformer and for various epochs (darker shades correspond to the last epochs). 
For the model trained with layerscale, the norm of the residual branch is on average 20\% of the norm of the main branch. 
We observe that the contribution of the residual blocks fluctuates more for the model trained without layerscale and in particular is lower for some of the deeper layers.  %
\label{fig:layerscale_analysis}}
\end{figure}

\paragraph{Statistics of branch weighting.}  
We evaluate the impact of Layerscale for a 36-blocks transformer by measuring the ratio between the norm of the residual activations and the norm of the activations of the main branch $ \| g_l(x) \|_2 / \| x \|_2 $.
The results are shown in Figure~\ref{fig:layerscale_analysis}.
We can see that training a model with Layerscale makes this ratio more uniform across layers, and seems to prevent some layers from having a disproportionate impact on the activations. Similar to prior works~\cite{Bachlechner2020ReZeroIA,Zhang2019FixupIR} we hypothetize that the benefit is mostly the impact on optimization. This hypothesis is supported by the control experiment that we detail in Appendix~\ref{sec:variant_layerscale}. 

\subsection{Class-attention layers} 
\label{sec:exp_class_attention}

\begin{table}
    \caption{Variations on CLS with Deit-Small (no LayerScale): we change the layer at which the class embedding is inserted. In ViT and DeiT, it is inserted at layer 0 jointly with the projected patches. We evaluate a late insertion of the class embedding, as well as our design choice to introduce specific class-attention layers. 
    \label{tab:ablation_pos_layer_insertion}}
    \centering
    \scalebox{0.9}{
    \begin{tabular}{cccccc}
    \toprule                                    
    depth: SA+CA & insertion layer & top-1 acc. & \#params & FLOPs\\ 
    \midrule
    \multicolumn{5}{c}{Baselines: DeiT-S and average pooling} \\ 
    \arrayrulecolor{black!30}\midrule \arrayrulecolor{black}                                     
    12: 12 + 0 & 0      & 79.9 & 22M & 4.6B\\ 
    12: 12 + 0 & n/a    & 80.3 & 22M & 4.6B\\ 
    \midrule
    \multicolumn{5}{c}{Late insertion of class embedding} \\ 
    \arrayrulecolor{black!30}\midrule \arrayrulecolor{black}                                     
    12: 12 + 0 & 2      & 80.0 & 22M & 4.6B\\ 
    12: 12 + 0 & 4      & 80.0 & 22M & 4.6B\\ 
    12: 12 + 0 & 8      & 80.0 & 22M & 4.6B\\ 
    12: 12 + 0 & 10     & 80.5 & 22M & 4.6B\\ 
    12: 12 + 0 & 11     & 80.3 & 22M & 4.6B\\ 
    \midrule                                     
    \multicolumn{5}{c}{DeiT-S with class-attention stage (SA+FFN)} \\ 
    \arrayrulecolor{black!30}\midrule \arrayrulecolor{black}                                     
    12: \pzo9 + 3  & 9      & 79.6 & 22M & 3.6B \\
    12: 10 + 2 & 10     & 80.3 & 22M & 4.0B\\
    12: 11 + 1 & 11     & 80.6 & 22M & 4.3B\\
    \arrayrulecolor{black!30}\midrule \arrayrulecolor{black}                                     
    13: 12 + 1 & 12     & 80.8 &  24M   & 4.7B\\
    14: 12 + 2 & 12     & 80.8 &  26M & 4.7B\\
    15: 12 + 3 & 12     & 80.6 &  27M  & 4.8B\\
    \bottomrule
    \end{tabular}}
\end{table}

In Table~\ref{tab:ablation_pos_layer_insertion} we study the impact on performance of the design choices related to class embedding. We depict some of them in Figure~\ref{fig:mix_and_summarize}. 
As a baseline, average pooling of patches embeddings with a vanilla DeiT-Small  achieves a better performance than using a class token. This choice, which does not employ any class embedding, is typical in convolutional networks, but possibly weaker with transformers when transferring to other tasks~\cite{el2021training}.  

\paragraph{Late insertion.} 
The performance increases when we insert the class embedding later in the transformer. It is maximized two layers before the output. 
Our interpretation is that the attention process is less perturbed in the 10 first layers, yet it is best to keep 2 layers for compiling the patches embedding into the class embedding via class-attention, otherwise the processing gets closer to a weighted average. 

\paragraph{Our class-attention layers} are designed on the assumption that there is no benefit in copying information from the class embedding back to the patch embeddings in the forward pass. Table~\ref{tab:ablation_pos_layer_insertion} supports that hypothesis: 
if we compare the performance for a total number of layers fixed to 12, the performance of \OURS with 10 SA and 2 CA layers is identical to average pooling and better than the DeiT-Small baseline with a lower number of FLOPs.  
If we set 12 layers in the self-attention stage, which dominates the complexity, we increase the performance significantly by adding two blocks of CA+FFN.

\subsection{Our \OURS models}
\label{sec:models}

\begin{table*}

    \caption{\OURS models: The design parameters are depth and $d$. The mem columns correspond to the memory usage. All models are initially trained at resolution 224 during 400 epochs. We also fine-tune these models at resolution 384 (identified by $\uparrow$384) or train them with distillation (\alambic). The FLOPs are reported for each resolution. 
    \label{tab:architectures}}  
    \scalebox{0.9}{
    \begin{tabular}{l|c@{\ \ }c@{\ \ }c|cc|cc|cc}
    \toprule
         CAIT & depth & $d$ & \#params & \multicolumn{2}{|c|}{FLOPs ($\times10^9$)} & \multicolumn{4}{c}{Top-1 acc. (\%): Imagenet1k-val}   \\

         model & (SA+CA) & &  ($\times10^6$)  & @224 & @384 & @224 &  $\uparrow$384 & @224\alambic & $\uparrow$384\alambic \\ 

        \midrule 
        XXS-24    & 24\,+\,2 & 192 & \pzo12.0 & \dzo2.5  & \dzo9.5  & 77.6 & 80.4 & 78.4 & 80.9    \\
        XXS-36    & 36\,+\,2 & 192 & \pzo17.3 & \dzo3.8  & \pzo14.2 & 79.1 & 81.8     & 79.7 & 82.2    \\ 
        \mycolrule                                                                                               
        XS-24     & 24\,+\,2 & 288 & \pzo26.6 & \dzo5.4  & \pzo19.3 & 81.8 & 83.8     & 82.0 &  84.1       \\
        XS-36     & 36\,+\,2 & 288 & \pzo38.6 & \dzo8.1  & \pzo28.8 & 82.6 & 84.3  & 82.9     &     84.8 \\
        \mycolrule                                                                                               
        S-24      & 24\,+\,2 & 384 & \pzo46.9 & \dzo9.4  & \pzo32.2 & 82.7 & 84.3 & 83.5 & 85.1    \\ 
        S-36      & 36\,+\,2 & 384 & \pzo68.2 & \pzo13.9 & \pzo48.0 & 83.3 & 85.0 & 84.0 & 85.4    \\
        S-48      & 48\,+\,2 & 384 & \pzo89.5 & \pzo18.6 & \pzo63.8 & 83.5 & 85.1 & 83.9 & 85.3    \\
        \mycolrule                                                                                               
        M-24      & 24\,+\,2 & 768 & 185.9    & \pzo36.0 & 116.1    & 83.4 & 84.5 & 84.7 & 85.8    \\
        M-36      & 36\,+\,2 & 768 & 270.9    & \pzo53.7 & 173.3    & 83.8 & 84.9 & 85.1 & 86.1    \\ 

    \bottomrule     
 \end{tabular}     
 } % scalebox 
%\bigskip

\end{table*}

\iffalse
\begin{table*}
\caption{Hyper-parameters for training \OURS models: The only parameters that we adjust per model are the drop rate $d_r$ of stochastic depth and the  LayerScale initialization~$\varepsilon$. 
\label{tab:hparams}}
    \centering \scalebox{0.9}{
    \begin{tabular}{l|cc}
    \toprule
         CAIT & \multicolumn{2}{c}{hparams} \\
         model & $d_r$ & $\varepsilon$   \\ 
        \midrule 
        XXS-24    & 0.05    & 10$^{-5}$  \\
        XXS-36    & 0.1\pzo & 10$^{-6}$  \\  
        \mycolrule                  
        XS-24     & 0.05    & 10$^{-5}$  \\
        XS-36     & 0.1\pzo & 10$^{-6}$  \\
        \mycolrule                  
        S-24      & 0.1\pzo & 10$^{-5}$  \\ 
        S-36      & 0.2\pzo & 10$^{-6}$  \\
        S-48      & 0.3\pzo & 10$^{-6}$  \\
        \mycolrule                  
        M-24      & 0.2\pzo & 10$^{-5}$  \\
        M-36      & 0.3\pzo & 10$^{-6}$  \\ 

    \bottomrule     
 \end{tabular}     
 } 
\end{table*}
\fi

\begin{table*}
\def \mysp {}
\caption{Hyper-parameters for training \OURS models: The only parameters that we adjust per model are the drop rate $d_r$ of stochastic depth and the  LayerScale initialization~$\varepsilon$. 
\label{tab:hparams}}
    \centering \scalebox{0.8}{
    \begin{tabular}{@{\ }l@{\hspace{6pt}}l|ccccccccccc@{\ }}
    \toprule
        \multicolumn{2}{l|}{CAIT model} & XXS-24 & XXS-36 & XS-24 & XS-36 & S-24 & S-36 & S-48 & M-24 & M-36 & M-48 \\
          \multirow{2}{*}{hparams} & $d_r$           &  0.05    & 0.1 & 0.05    & 0.1\pzo & 0.1\pzo &  0.2\pzo & 0.3\pzo & 0.2\pzo & 0.3\pzo & 0.4\pzo   \\
          & $\varepsilon$   & 10$^{-5}$ & 10$^{-6}$ & 10$^{-5}$ & 10$^{-6}$ & 10$^{-5}$ & 10$^{-6}$& 10$^{-6}$&  10$^{-5}$ & 10$^{-6}$ & 10$^{-6}$    \\
    \bottomrule     
 \end{tabular}} 
\end{table*}

Our \OURS models are built upon ViT: the only difference is that we incorporate LayerScale in  each residual block (see Section~\ref{sec:training}) and the two-stages architecture with class-attention layers described in Section~\ref{sec:method}.    
Table \ref{tab:architectures} describes our different models. 
The design parameters governing the capacity are the depth and the working dimensionality $d$. In our case $d$ is related to the number of heads $h$ as $d=48\times h$, since we fix the \textbf{number of components per head} to 48. This choice is a bit smaller than the value used in DeiT. We also adopt the \textbf{crop-ratio} of 1.0 optimized for DeiT by Wightman~\cite{pytorchmodels}. 
Table~\ref{tab:ablation_nb_heads} and~\ref{tab:crop_ratio} in the ablation section~\ref{sec:ablation} support these choices. 

We incorporate talking-heads attention~\cite{Shazeer2020TalkingHeadsA} into our model. It increases the performance on Imagenet of DeiT-Small from 79.9\% to 80.3\%.

\paragraph{The hyper-parameters} are identical to those provided in DeiT~\cite{Touvron2020TrainingDI}, except mentioned otherwise. We use a batch size of 1024 samples and train during 400 epochs with repeated augmentation~\cite{berman2019multigrain,hoffer2020augment}. The learning rate of the AdamW optimizer~\cite{Loshchilov2017AdamW} is set to 0.001 and associated with a cosine training schedule, 5 epochs of warmup and a weight decay of 0.05. 
We report in Table~\ref{tab:hparams} the two hyper-parameters that we modify depending on the model complexity, namely the drop rate $d_r$ associated with uniform stochastic depth, and the initialization value $\varepsilon$ of LayerScale.  

\paragraph{Fine-tuning at higher resolution ($\uparrow$) and distillation (\alambicb).} 
We train all our models at resolution 224, and optionally fine-tune them at a higher resolution to trade performance against accuracy~\cite{dosovitskiy2020image,Touvron2020TrainingDI,Touvron2019FixRes}: we denote the model by $\uparrow$384 models fine-tuned at resolution 384$\times$384. 
We also train models with distillation (\alambicb) as suggested by Touvron~\etal.~\cite{Touvron2020TrainingDI}. We use a RegNet-16GF~\cite{Radosavovic2020RegNet} as teacher and adopt the ``hard distillation''~\cite{Touvron2020TrainingDI} for its simplicity.

\subsection{Results}
\label{sec:results}

\subsubsection{Performance/complexity of \OURS models}

Table~\ref{tab:architectures} provides different complexity measures for our models. As a general observation, we observe a subtle interplay between the width and the depth, both contribute to the performance as reported by Dosovitskiy \etal. \cite{dosovitskiy2020image} with longer training schedules. But if one parameter is too small the gain brought by increasing the other is not worth the additional complexity.  

Fine-tuning to size 384 ($\uparrow$) systematically offers a large boost in performance without changing the number of parameters. 
It also comes with a higher computational cost. 
In contrast, leveraging a pre-trained convnet teacher with hard distillation as suggested by Touvron \etal~\cite{Touvron2020TrainingDI} provides a boost in accuracy without affecting the number of parameters nor the speed. 

\begin{table}
    %\centering
    \caption{\textbf{Complexity vs accuracy} on Imagenet~\cite{Russakovsky2015ImageNet12}, Imagenet Real~\cite{Beyer2020ImageNetReal} and Imagenet V2 matched frequency~\cite{Recht2019ImageNetv2} for models trained without external data. We compare \ours with DeiT~\cite{Touvron2020TrainingDI}, Vit-B~\cite{dosovitskiy2020image}, TNT~\cite{Han2021TransformerIT}, T2T~\cite{Yuan2021TokenstoTokenVT} and to several state-of-the-art convnets:   Regnet~\cite{Radosavovic2020RegNet} improved by Touvron et al.~\cite{Touvron2020TrainingDI}, EfficientNet~\cite{Cubuk2019RandAugmentPA,tan2019efficientnet,Xie2020AdversarialEI}, Fix-EfficientNet~\cite{Touvron2020FixingTT} and NFNets~\cite{Brock2021HighPerformanceLI}. Most reported results are from corresponding papers, and therefore the training procedure differs for the different models. For Imagenet V2 matched frequency and Imagenet Real we report the results provided by the authors. When not available (like NFNet), we report the results measured by Wigthman~\cite{pytorchmodels} with converted models, which may be suboptimal. The RegNetY-16GF is the teacher model that we trained for distillation. We report the best result in \textbf{bold} and the second best result(s) \underline{underlined}.  \label{tab:throughput}} 
    \def \mysp {\hspace{7pt}}
    \centering \scalebox{0.87}
    {%\small 
    \begin{tabular}{@{\ }lr@{\mysp}r@{\mysp}|cc|c|c|c@{\ }}
    \toprule
             &   nb of \ & nb of \pzo     &     \multicolumn{2}{c|}{image size}  & \multicolumn{1}{c}{ImNet} & \multicolumn{1}{|c|}{Real}& \multicolumn{1}{c}{V2}\\ 
    Network  & param. & FLOPs & train & test   &  top-1  &  top-1 &  top-1 \\
    \toprule
    RegNetY-16GF       & \pzo84M & \pzo16.0B & $224$ & $224$ & 82.9 & 88.1 & 72.4 \\
    \midrule

    EfficientNet-B5    & \pzo30M & \dzo9.9B  & $456$ & $456$  & 83.6 & 88.3 & 73.6 \\
    EfficientNet-B7    & \pzo66M & \pzo37.0B & $600$ & $600$  & 84.3 & \_   & \_   \\
    \midrule   
   EfficientNet-B5 RA & \pzo30M & \dzo9.9B  & $456$ & $456$ & 83.7 & \_   & \_ \\
   EfficientNet-B7 RA & \pzo66M & \pzo37.0B & $600$ & $600$  & 84.7 & \_   & \_ \\
   \midrule
   EfficientNet-B7 AdvProp \hspace{-15pt} & \pzo66M & \pzo37.0B & $600$ & $600$  & 85.2 & 89.4   & 76.0 \\
    \midrule
    Fix-EfficientNet-B8 & \pzo87M & 89.5B & $672$ & $800$  & 85.7 & \underline{90.0}   & 75.9 \\
    \midrule
    NFNet-F0           & \pzo72M & \pzo12.4B & $192$ & $256$ & 83.6 & 88.1 & 72.6 \\
    NFNet-F1           & 133M    & \pzo35.5B & $224$ & $320$  & 84.7 & 88.9 & 74.4 \\
    NFNet-F2           & 194M    & \pzo62.6B & $256$ & $352$  & 85.1 & 88.9 & 74.3 \\
    NFNet-F3           & 255M    & 114.8B    & $320$ & $416$  & 85.7 & 89.4 & 75.2 \\
    NFNet-F4           & 316M    & 215.3B    & $384$ & $512$   & 85.9 & 89.4 & 75.2 \\
    NFNet-F5           & 377M    & 289.8B    & $416$ & $544$   & 86.0 & 89.2 & 74.6 \\
    NFNet-F6+SAM       & 438M    & 377.3B    & $448$ & $576$   & \textbf{86.5} & 89.9 & 75.8 \\
    \toprule
    \multicolumn{8}{c}{Transformers}\\
    \midrule
    ViT-B/16           & \pzo86M & \pzo55.4B & $24$ & $384$ & 77.9 & 83.6 & \_ \\
    ViT-L/16           & 307M    & 190.7B    & $224$ & $384$  & 76.5 & 82.2 & \_ \\
    \midrule
    T2T-ViT t-14       & \pzo21M & \dzo5.2B  & $224$ & $224$        & 80.7 & \_   & \_ \\
    \midrule
    TNT-S              & \pzo24M & \dzo5.2B  & $224$ & $224$       & 81.3 & \_   & \_ \\
    TNT-S + SE         & \pzo25M & \dzo5.2B  & $224$ & $224$        & 81.6 & \_   & \_  \\
    TNT-B              & \pzo66M & \pzo14.1B & $224$ & $224$       & 82.8 & \_   & \_ \\
    \midrule
    DeiT-S             & \pzo22M & \dzo4.6B  & $224$ & $224$  & 79.8  & 85.7  & 68.5 \\
    DeiT-B             & \pzo86M & \pzo17.5B & $224$ & $224$ &  81.8 &   86.7 & 71.5\\
    DeiT-B$\uparrow$384         & \pzo86M & \pzo55.4B & $224$  &  $384$ &  83.1 & 87.7 & 72.4\\
    DeiT-B$\uparrow$384\alambicb 1000 epochs \hspace{-15pt} & 87M  & 55.5B & $224$ &  $384$ &  85.2   & 89.3 & 75.2 \\

    \toprule
    \multicolumn{8}{c}{Our deep transformers} \\
    \midrule
    \OURS-S36   & \pzo68M & \pzo13.9B &  $224$ &  $224$ &  83.3 & 88.0 & 72.5\\
    \OURS-S36$\uparrow$384 & \pzo68M & \pzo48.0B &  $224$ & $384$ &  85.0 & 89.2  & 75.0 \\
    \OURS-S48$\uparrow$384 & \pzo89M & \pzo63.8B &  $224$ & $384$ &  85.1 & 89.5  & 75.5 \\
    \midrule
    \OURS-S36\alambicb& \pzo68M & \pzo13.9B &  $224$ &  $224$  &  84.0 & 88.9 & 74.1\\
    \OURS-S36$\uparrow$384\alambicb  & \pzo68M & \pzo48.0B &  $224$ &  $384$ &  85.4 & 89.8 & 76.2 \\
%    \OURS-M36\alambicb & 271M & \pzo53.7B &  $224$ &  $224$ &  85.1 & 89.3 & 74.9 \\
    \OURS-M36$\uparrow$384\alambicb & 271M & 173.3B &   $224$   & $384$ &  86.1 & \underline{90.0} & 76.3 \\
    \OURS-M36$\uparrow$448\alambicb & 271M & 247.8B & $224$ & $448$  & \underline{86.3} & \textbf{90.2} & \underline{76.7}\\
    \OURS-M48$\uparrow$448\alambicb & 356M & 329.6B & $224$ & $448$  & \textbf{86.5} & \textbf{90.2} & \textbf{76.9}\\
    \bottomrule
    \end{tabular}}
\end{table}

\subsubsection{Comparison with the state of the art on Imagenet}  

Our main classification experiments are carried out on ImageNet~\cite{Russakovsky2015ImageNet12}, and also evaluated on two variations of this dataset: ImageNet-Real~\cite{Beyer2020ImageNetReal} that corrects and give a more detailed annotation, and ImageNet-V2~\cite{Recht2019ImageNetv2} (matched frequency) that provides a separate test set. 
In Table~\ref{tab:throughput} we compare some of our models with the state of the art on Imagenet classification when training without external data. We focus on the models \OURS-S36 and \OURS-M36, at different resolutions and with or without distillation. 

On Imagenet1k-val, \OURS-M48$\uparrow$448\alambicb  achieves \bestperf of top-1 accuracy, which is a significant improvement over DeiT (85.2\%). It is the state of the art, on par with a recent concurrent work \cite{Brock2021HighPerformanceLI} that has a significantly higher number of FLOPs. 
%that reports a top-1 accuracy of 86.5\%. 
%Our \OURS-M36$\uparrow$384\alambicb model obtains 86.1\% top-1 accuracy on Imagenet-1k-val, while: this network is more efficient that the best performing NFNet convnets w.r.t. FLOPs and even more in terms of throughput (images processed per second on a V100 GPU), thanks to the lower memory usage. 
%
Our approach outperforms the state of the art on Imagenet with reassessed labels, and on Imagenet-V2, which has a distinct validation set which makes it harder to overfit. % 

\begin{figure}[t]
%\framebox{
\begin{minipage}{0.99\linewidth}
\includegraphics[trim = 0 12 0 0, clip, height=0.48\linewidth]{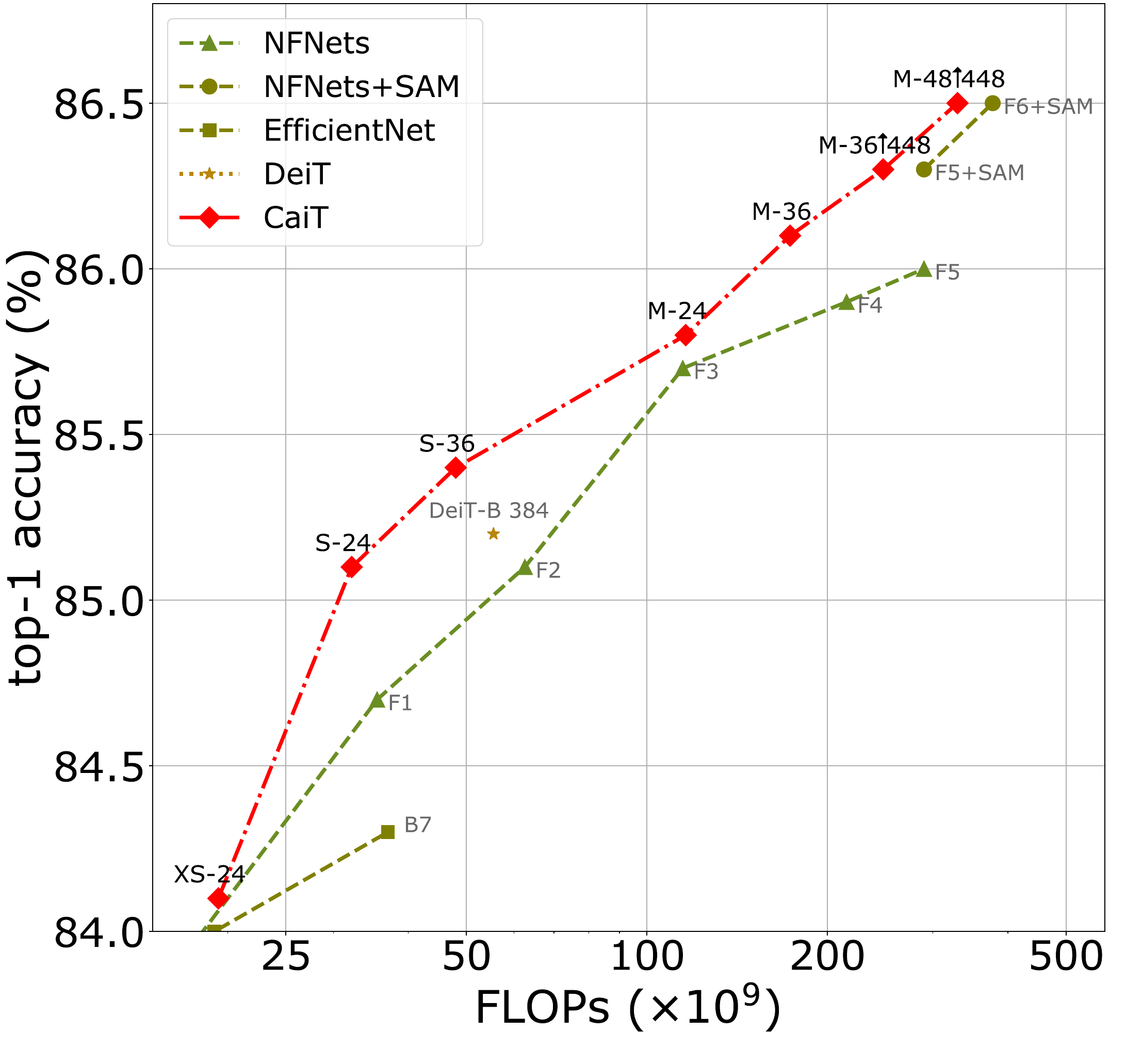}
\hfill
\includegraphics[trim = 90 12 10 0, clip, height=0.48\linewidth]{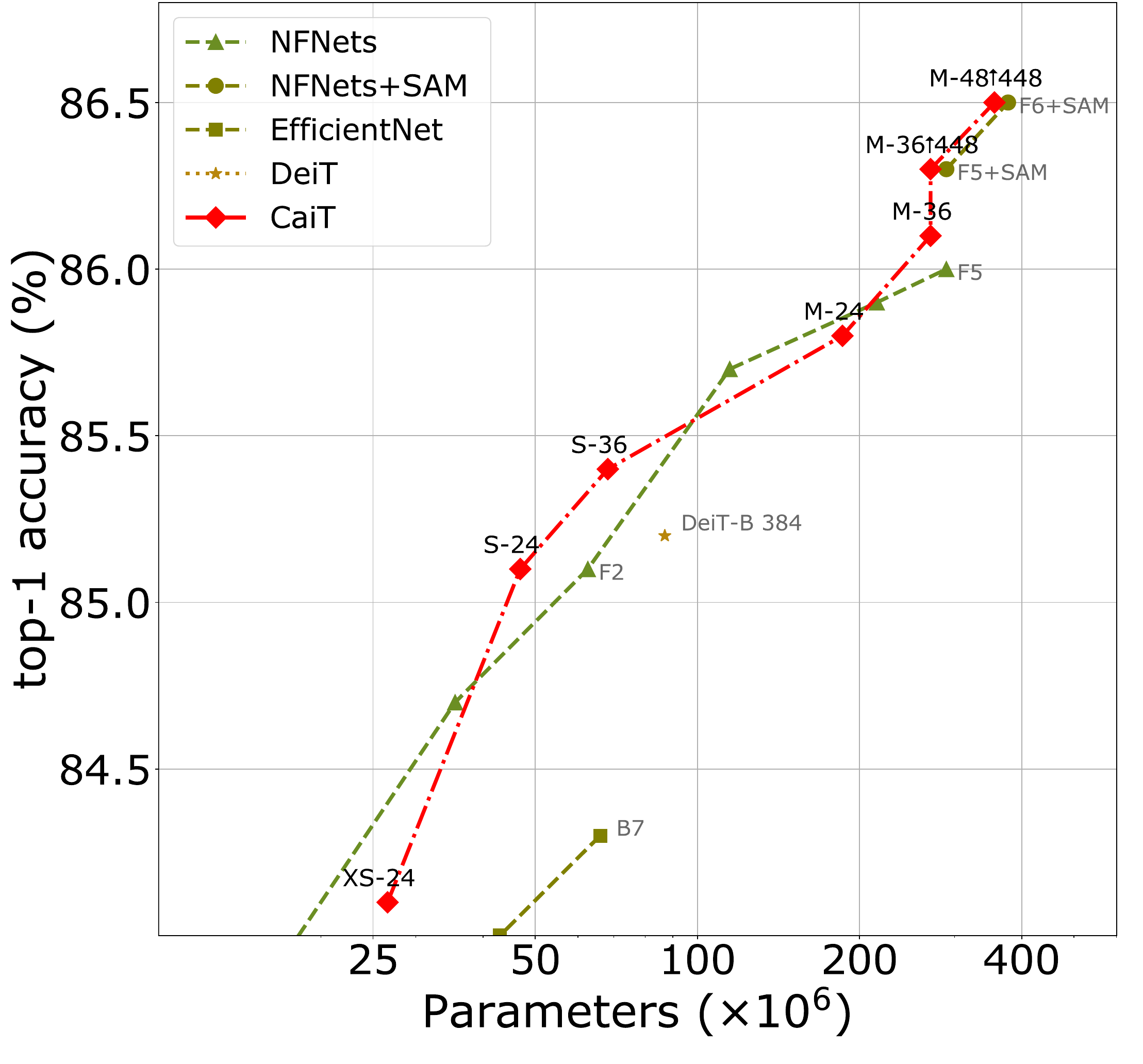}
\end{minipage}%    }

%\end{frame}
%
\caption{We represent FLOPs and parameters for our best \ours$\uparrow384$ and $\uparrow448$ \alambic models trained with distillation. They are competitive on ImageNet-1k-val with the sota in the high accuracy regime, from XS-24 to M-48. Convolution-based neural networks like NFNets and EfficientNet are better in low-FLOPS and low-parameters regimes. 
\label{fig:comp_flops}}
\end{figure}

%------
\subsubsection{Transfer learning}
\label{sec:transfer_learning}

We evaluated our method on transfer learning  tasks by fine-tuning on the datasets in Table~\ref{tab:dataset}.

\paragraph{Fine-tuning procedure. } 
For fine-tuning we use the same hyperparameters as for training.
We only decrease the learning rates by a factor 10 (for CARS, Flowers, iNaturalist), 100 (for CIFAR-100, CIFAR-10) and adapt the number of epochs (1000 for CIFAR-100, CIFAR-10, Flowers-102 and Cars-196, 360 for iNaturalist 2018 and 2019). We have not used distillation for this finetuning. 

\paragraph{Results.}
Table~\ref{tab:transfer} compares \ours transfer learning results  to those of  EfficientNet~\cite{tan2019efficientnet}, %
ViT~\cite{dosovitskiy2020image} and DeiT~\cite{Touvron2020TrainingDI}. 
These results show the excellent generalization of the transformers-based models in general. Our \OURS models achieve excellent results, as shown by the overall better performance than EfficientNet-B7 across datasets.  

\begin{table}[t]
\caption{Datasets used for our different tasks.  \label{tab:dataset}}
\centering
\scalebox{0.9}{
\begin{tabular}{l|rrr}
\toprule
Dataset & Train size & Test size & \#classes   \\
\midrule
ImageNet \cite{Russakovsky2015ImageNet12}  & 1,281,167 & 50,000 & 1000  \\ 
iNaturalist 2018~\cite{Horn2018INaturalist}& 437,513   & 24,426 & 8,142 \\ 
iNaturalist 2019~\cite{Horn2019INaturalist}& 265,240   & 3,003  & 1,010  \\ 
Flowers-102~\cite{Nilsback08}& 2,040 & 6,149 & 102  \\ 
Stanford Cars~\cite{Cars2013}& 8,144 & 8,041 & 196  \\  
CIFAR-100~\cite{Krizhevsky2009LearningML}  & 50,000    & 10,000 & 100   \\ 
CIFAR-10~\cite{Krizhevsky2009LearningML}  & 50,000    & 10,000 & 10   \\ 
\bottomrule
\end{tabular}}%}
\end{table}

\def \mysp {\hspace{9pt}}
\begin{table}
    \caption{Results in transfer learning. All models are trained and evaluated at resolution 224 and with a crop-ratio of 0.875 in this comparison (see Table~\ref{tab:crop_ratio} for the comparison of crop-ratio on Imagenet). 
    \label{tab:transfer}}
    \centering
    \scalebox{0.9}{
    \begin{tabular}{@{\ }l@{\mysp}|@{\mysp}c@{\mysp}|@{\mysp}c@{\mysp}c@{\mysp}c@{\mysp}c@{\mysp}c@{\mysp}c@{\mysp}c@{\ }}
    \toprule
    Model                                      
        & \rotatebox{90}{ImageNet}
        & \rotatebox{90}{CIFAR-10}
        & \rotatebox{90}{CIFAR-100}  
        & \rotatebox{90}{Flowers} 
        & \rotatebox{90}{Cars} 
        & \rotatebox{90}{iNat-18} 
        & \rotatebox{90}{iNat-19} 
        & \rotatebox{90}{FLOPs}\\
    \midrule

    EfficientNet-B7 & 84.3 & 98.9 & 91.7  & 98.8 & \textbf{94.7} & \_ & \_ & \pzo37.0B \\
    \mycolrule       
    ViT-B/16        & 77.9 & 98.1 & 87.1 & 89.5 & \_   & \_   & \_   &  \pzo55.5B\\
    ViT-L/16        & 76.5 & 97.9 & 86.4 & 89.7 & \_   & \_   & \_   &  190.7B\\
    \mycolrule         
    Deit-B  224     & 81.8 & 99.1 & 90.8 & 98.4 & 92.1 & 73.2 & 77.7 &  \pzo17.5B \\
    \midrule
    \ours-S-36 224  & 83.4 & 99.2 & 92.2 & 98.8 & 93.5 & 77.1 & 80.6 & \pzo13.9B\\
    \ours-M-36 224  & 83.7 & 99.3 & 93.3 & 99.0 & 93.5 & 76.9 & 81.7 & \pzo53.7B\\
    \mycolrule       
    \ours-S-36 \alambic 224 & 83.7 & 99.2 & 92.2 & 99.0 & 94.1  & 77.0 & 81.4 & \pzo13.9B\\
    \ours-M-36 \alambic 224 & \textbf{84.8} & \textbf{99.4} & \textbf{93.1}&  \textbf{99.1} & 94.2  & \textbf{78.0} & \textbf{81.8} & \pzo53.7B\\    
    \bottomrule
    \end{tabular}}
\end{table}

\subsection{Ablation} 
\label{sec:ablation}

In this section we provide different sets of ablation, in the form of a transition from DeiT to \OURS. Then we provide experiments that have guided our hyper-parameter optimization. As mentioned in the main paper, we use the same hyperparameters as in DeiT~\cite{Touvron2020TrainingDI} everywhere except stated otherwise. We have only changed the number of attention for a given working dimension (see Section \ref{sec:ablation_heads}), and changed the crop-ratio (see Section~\ref{sec:ablation_cropratio}).

\begin{table}
    \caption{Ablation: we present the ablation path from DeiT-S to our CaiT models. We highlight the complementarity of our approaches and optimized hyper-parameters. Note, Fine-tuning at higher resolution supersedes the inference at higher resolution.  See Table~\ref{tab:init_comp} for adapting stochastic depth before adding LayerScale. % \newline 
    $\dagger$: training failed. 
    \label{tab:ablation_small}} %
 \centering 
    \scalebox{0.9}{
    \begin{tabular}{lc@{\ \ }c@{\ \ }c}
    \toprule
         Improvement & top-1 acc. & \#params & FLOPs\\
         \midrule
         DeiT-S {\small [{\it d=384,300 epochs}]}     & 79.9 &  22M & \pzo4.6B  \\
         \midrule
         + More heads {\small [{\it 8}]}  & 80.0 &  22M & \pzo4.6B   \\
         + Talking-heads   & 80.5 &  22M & \pzo4.6B   \\
         + Depth {\small [{\it 36 blocks}]}     &  \ \,69.9$\dagger$        & 64M  & 13.8B\\
         \rowcolor{gray!10}
         \textbf{+ Layer-scale {\small [{\it init $\varepsilon=10^{-6}$}]}}          & 80.5 &  64M & 13.8B\\
         \rowcolor{gray!10}
         + Stch depth. adaptation {\small [{\it $d_r$=0.2}]}     &    83.0       &  64M& 13.8B\\
         \rowcolor{gray!10}
         \textbf{+ \OURS architecture} {\small [{\it specialized class-attention layers}]} & 83.2 &  68M      & 13.9B \\

         + Longer training {\small [{\it 400 epochs}]} & 83.4 &  68M      & 13.9B \\
         + Inference at higher resolution {\small [{\it 256}]} & 83.8 &  68M      & 18.6B \\
         + Fine-tuning at higher resolution {\small [{\it 384}]} \hspace{-15pt} & 84.8 &  68M      & 48.0B \\
         + Hard distillation {\small [{\it teacher: RegNetY-16GF}]} \hspace{-15pt} & 85.2 &  68M      & 48.0B \\
         + Adjust crop ratio {\small [{\it 0.875 $\rightarrow$ 1.0}]} \hspace{-15pt} & 85.4 &  68M      & 48.0B \\
         \bottomrule
    \end{tabular}}
\end{table}

\subsubsection{Step by step from DeiT-Small to \OURS-S36}
In Table~\ref{tab:ablation_small} we present how to gradually transform the  Deit-S~\cite{Touvron2020TrainingDI} architecture into \OURS-36, and measure at each step the performance/complexity changes. One can see that \OURS is complementary with LayerScale and offers an improvement without significantly increasing the FLOPs. 
As already reported in the literature, the resolution is another important step for improving the performance and fine-tuning instead of training the model from scratch saves a lot of computation at training time. Last but not least, our models benefit from longer training schedules.

\subsubsection{Optimization of the number of heads} 
\label{sec:ablation_heads}

In Table~\ref{tab:ablation_nb_heads} we study the impact of the number of heads for a fixed working dimensionality. 
This architectural parameter has an impact on both the accuracy, and the efficiency: while the number of FLOPs remain roughly the same, the compute is more fragmented when increasing this number of heads and on typical hardware this leads to a lower effective throughput. 
Choosing 8 heads in the self-attention offers a good compromise between accuracy and speed. In Deit-Small, this parameter was set to 6.  

\begin{table}[t]
    \caption{Deit-Small: for a fixed 384 working dimensionality and number of parameters, impact of the number of heads on the accuracy and throughput (images processed per second at inference time on a singe V100 GPU). %\rv{fused Cuda kernel for DeiT}
    \label{tab:ablation_nb_heads}}
    \centering
    \vspace{-0.5em}
    \scalebox{0.9}{
    \begin{tabular}{ccccc}
    \toprule
         \# heads & dim/head &  throughput (im/s) & GFLOPs & top-1 acc.\\
          
         \midrule
                   \pzo1 & 384     & 1079      & 4.6 & 76.80 \\ 
                   \pzo2 & 192     & 1056      & 4.6 & 78.06 \\ 
                   \pzo3 & 128     & 1043      & 4.6 & 79.35 \\ 
                   \pzo6 & \pzo64  & \pzo989   & 4.6 & 79.90 \\ 
\rowcolor{gray!10} \pzo8 & \pzo48  & \pzo971   & 4.6 & 80.02 \\ 
                   12    & \pzo32  & \pzo927   & 4.6 & 80.08 \\ 
                   16    & \pzo24  & \pzo860   & 4.6 & 80.04 \\
                   24    & \pzo16  & \pzo763   & 4.6 & 79.60 \\
          \bottomrule
    \end{tabular}}
\end{table}

\subsubsection{Adaptation of the crop-ratio}
\label{sec:ablation_cropratio}

In the typical (``center-crop'') evaluation setting, most convolutional neural networks crop a subimage with a given ratio, typically extracting a $224 \times 224$ center crop from a $256\times256$ resized image, leading to the typical ratio of 0.875. Wightman \etal~\cite{pytorchmodels} notice that setting this crop ratio to 1.0 for transformer models has a positive impact: the distilled DeiT-B$\uparrow 384$ reaches a top1-accuracy on Imagenet1k-val of $85.42\%$ in this setting, which is a gain of +0.2\% compared to the accuracy of 85.2\% reported by Touvron \etal~\cite{Touvron2020TrainingDI}. 

Our measurements concur with this observation: We observe a gain for almost all our models and most of the evaluation benchmarks. For instance our  model M36$\uparrow$384\alambicb increases to 86.1\% top-1 accuracy on Imagenet-val1k. 
\begin{table}[t]
    \caption{We compare performance with the defaut crop-ratio of 0.875 usually used with convnets, and the crop-ratio of 1.0~\cite{pytorchmodels} that we adopt for \OURS. 
    \label{tab:crop_ratio}}
    \centering
    \scalebox{0.9}{
    \begin{tabular}{c|cc|c|c|c}
    \toprule
     \multirow{2}{4em}{Network}   & \multicolumn{2}{c|}{Crop Ratio} & ImNet & Real  & V2 \\
                 \cmidrule{2-6}
                                  &  0.875 & 1.0                  & top-1 & top-1 & top-1 \\
    \midrule
    
    \multirow{2}{5em}{S36} 
                    &\checkmark & \_             & 83.4 & 88.1 & 73.0\\
                    &\_         & \checkmark     & 83.3 & 88.0 & 72.5 \\ 
    
    \midrule
    \multirow{2}{5em}{S36$\uparrow$384} 
                        &\checkmark & \_             & 84.8 & 88.9 & 74.7\\
                        &\_         & \checkmark     & 85.0 & 89.2 & 75.0 \\ 
                            
      \midrule                        
                            
         \multirow{2}{5em}{S36\alambicb} 
                            &\checkmark & \_             & 83.7 & 88.9 & 74.1\\
                            &\_         & \checkmark     & 84.0 & 88.9 & 74.1\\ 
                            
    \midrule     
        
        \multirow{2}{5em}{M36\alambicb} 
                            &\checkmark & \_             & 84.8 & 89.2 & 74.9\\
                            &\_         & \checkmark     & 84.9 & 89.2 & 75.0\\ 
                            
   \midrule                         
        \multirow{2}{5em}{S36$\uparrow$384\alambicb} 
                            &\checkmark & \_             & 85.2 & 89.7 & 75.7\\
                            &\_         & \checkmark     & 85.4 & 89.8 & 76.2 \\ 
    
    \midrule
    \multirow{2}{5em}{M36$\uparrow$384\alambicb} 
                            &\checkmark & \_             & 85.9 & 89.9 & 76.1\\
                            &\_         & \checkmark     & 86.1 & 90.0  & 76.3 \\

    \bottomrule
    \end{tabular}}
\end{table}

\subsubsection{Longer training schedules}

As shown in Table~\ref{tab:ablation_small}
, increasing the number of training epochs from 300 to 400 improves the performance of \ours-S-36. 
However, increasing the number of training epochs from 400 to 500 does not change performance significantly (83.44 with 400 epochs 83.42 with 500 epochs). 
This is consistent with the observation of the DeiT~\cite{Touvron2020TrainingDI} paper, which notes a saturation of performance from 400 epochs for the models trained without distillation.

\section{Visualizations}
\label{sec:vizualizations}

\def \mywi {0.19\linewidth}
\renewcommand \myfig[1] {\includegraphics[width=\mywi,height=\mywi]{figs/images/#1}}
\renewcommand \myfigc[1] {\includegraphics[trim=7 6 7 7,clip,width=\mywi,height=\mywi]{figs/images/#1}}
\begin{figure}
\scalebox{1.0}{
\begin{minipage}[r]{\linewidth}
          \hfill 
          \makebox[\mywi]{\scriptsize Head 1 $\downarrow$}  
          \makebox[\mywi]{\scriptsize Head 2 $\downarrow$}  
          \makebox[\mywi]{\scriptsize Head 3 $\downarrow$}  
          \makebox[\mywi]{\scriptsize Head 4 $\downarrow$}  \\[-1em]
          
          \myfigc{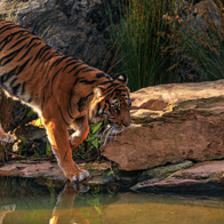} \hfill 
          \myfigc{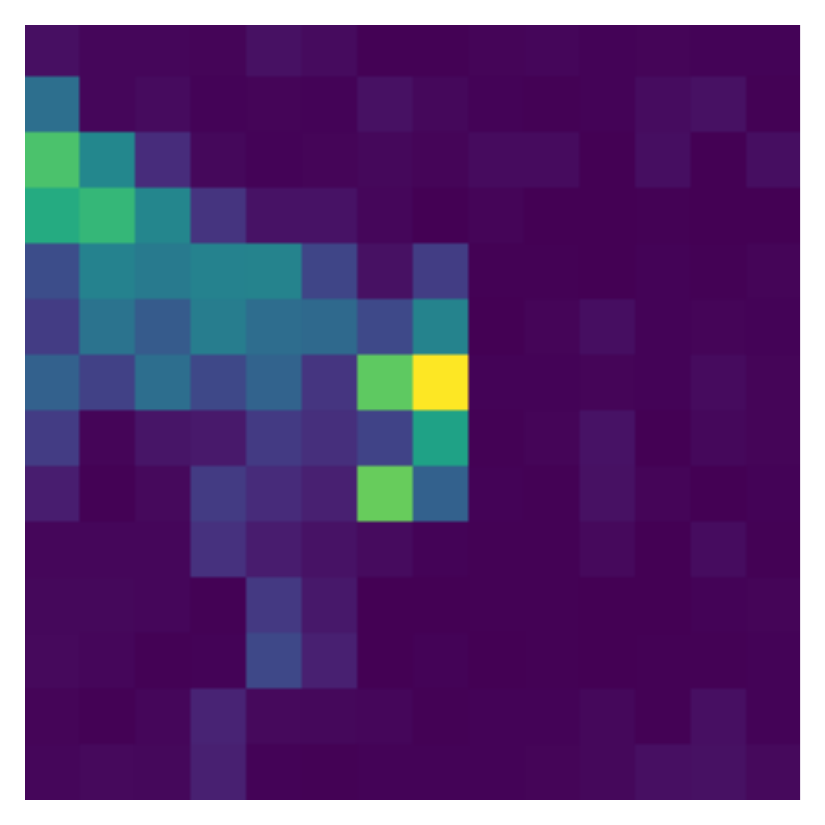} %
          \myfigc{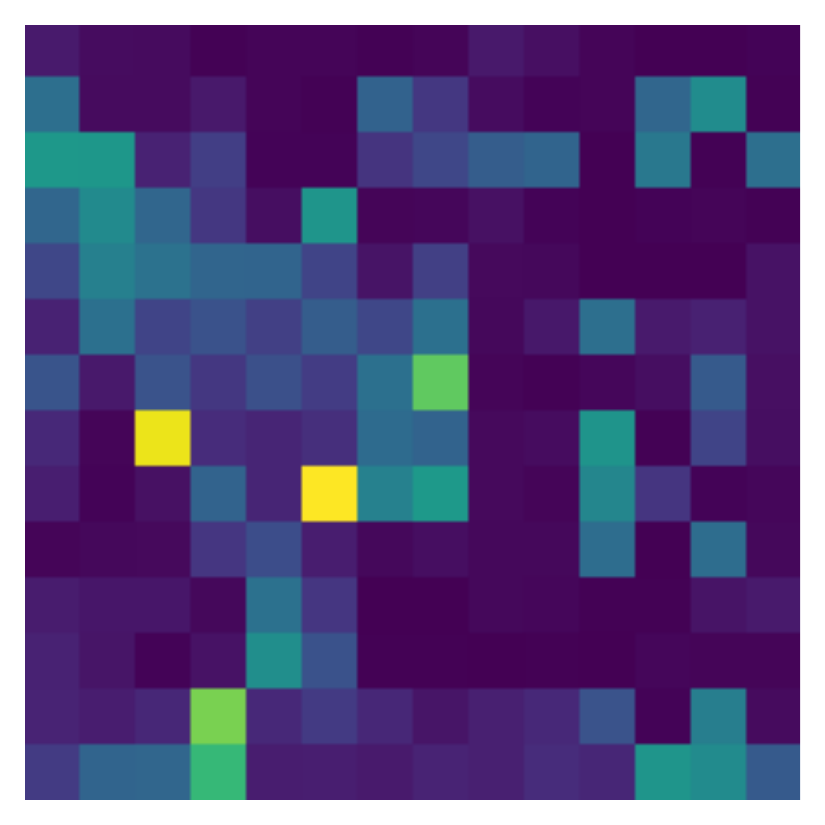} %
          \myfigc{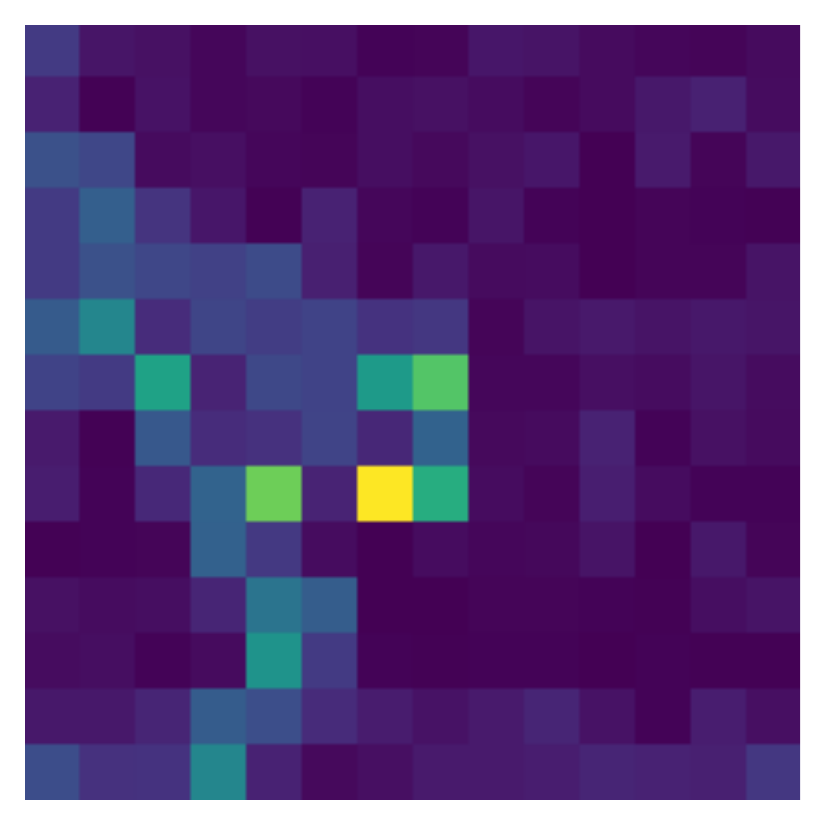} %
          \myfigc{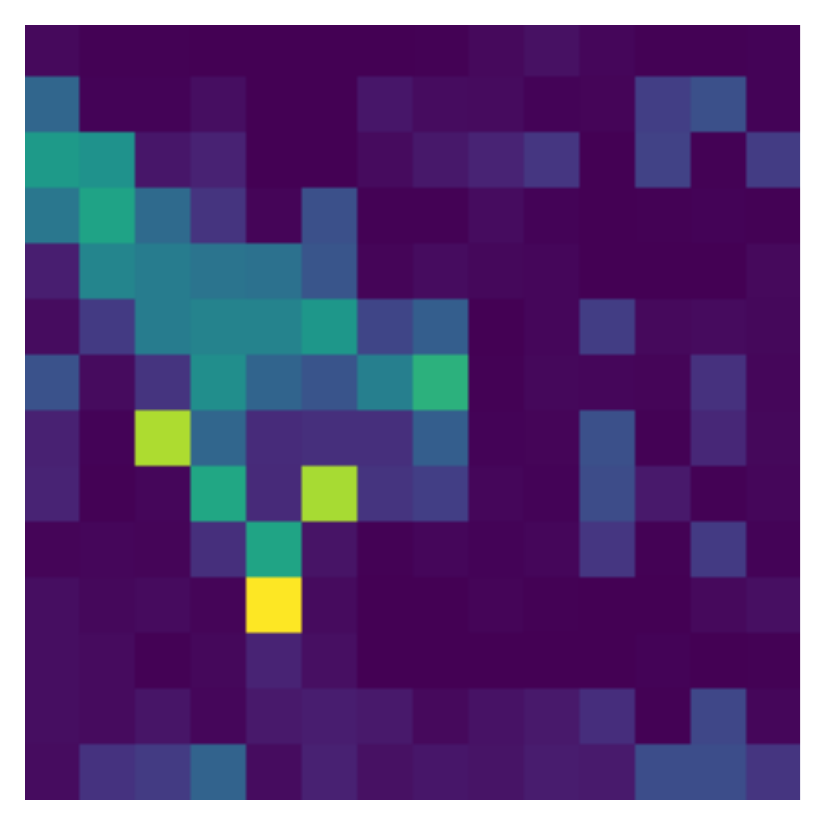} \\[-1em]
          
          ~\hspace{0.19\linewidth} \hfill
          \myfigc{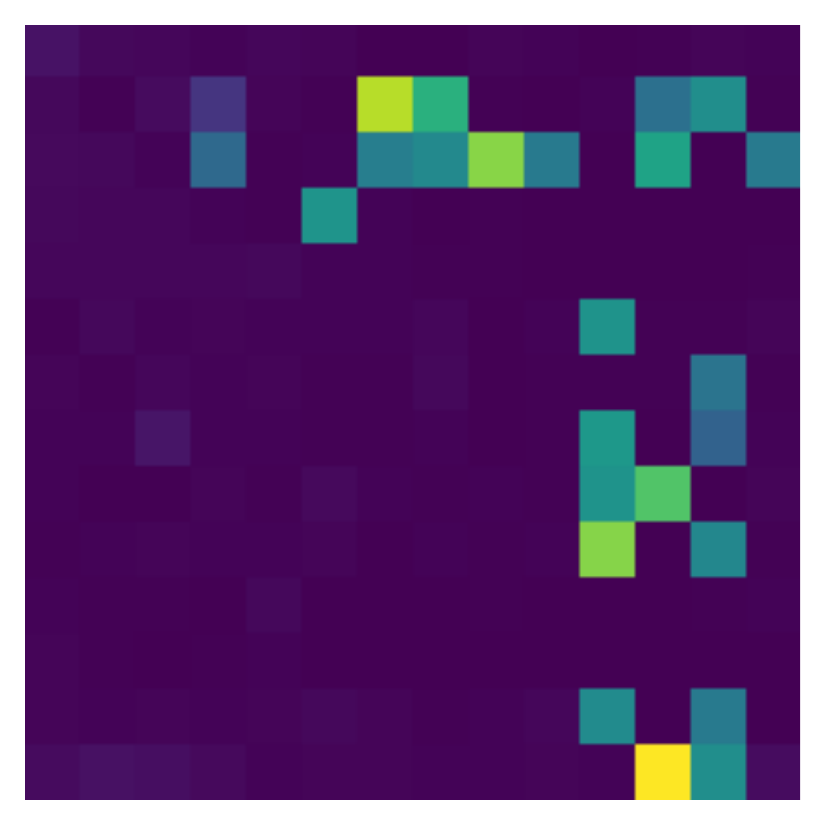} 
          \myfigc{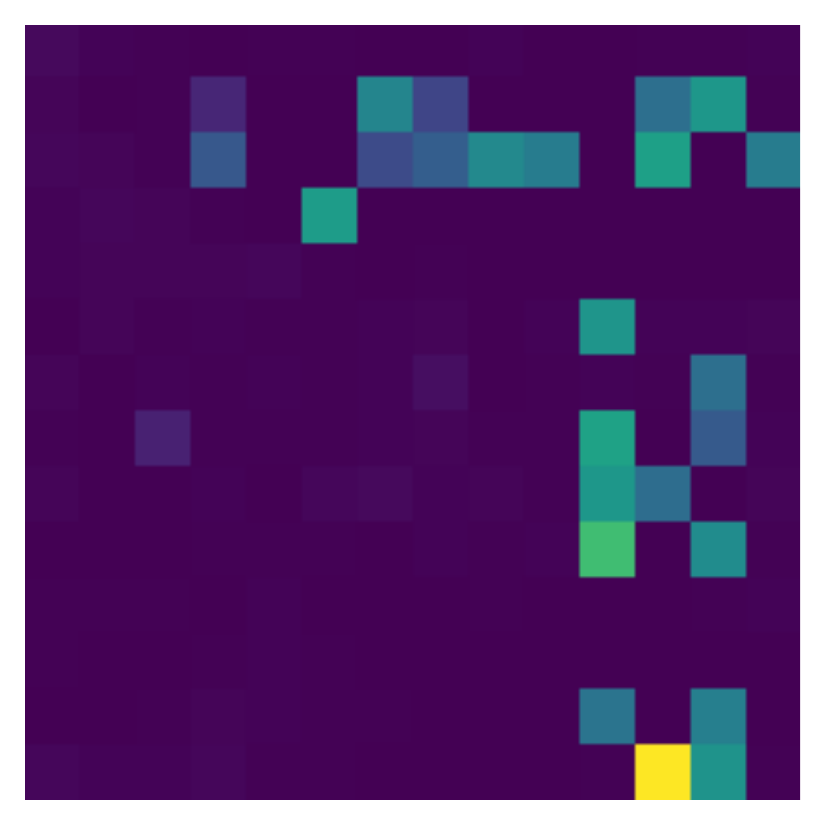} 
          \myfigc{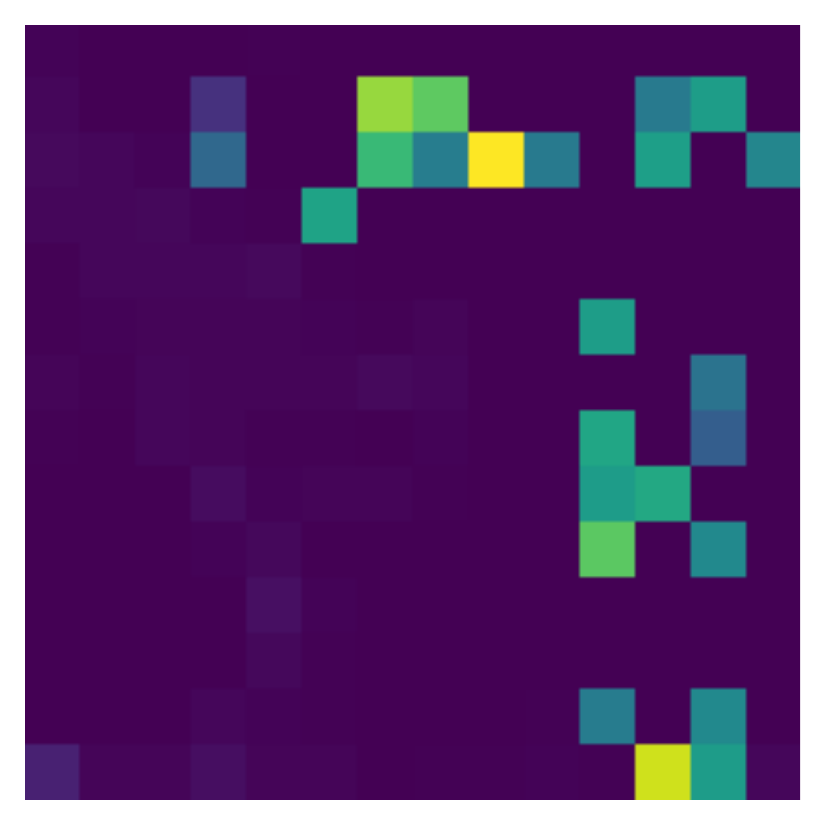} 
          \myfigc{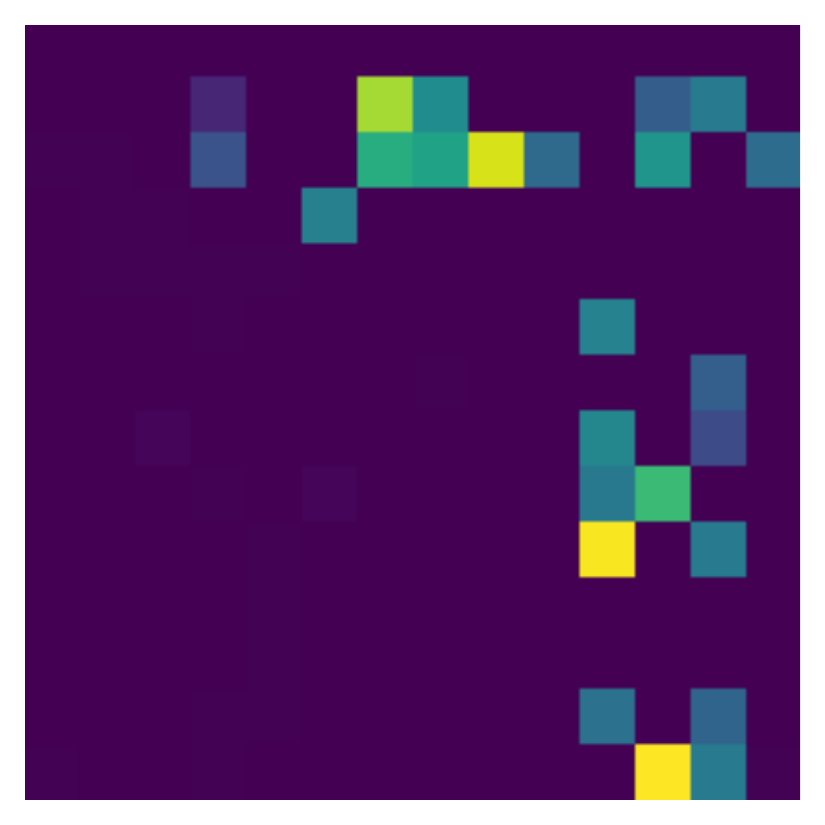} \\[-1em]
          
          \myfigc{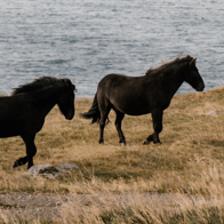} \hfill
          \myfigc{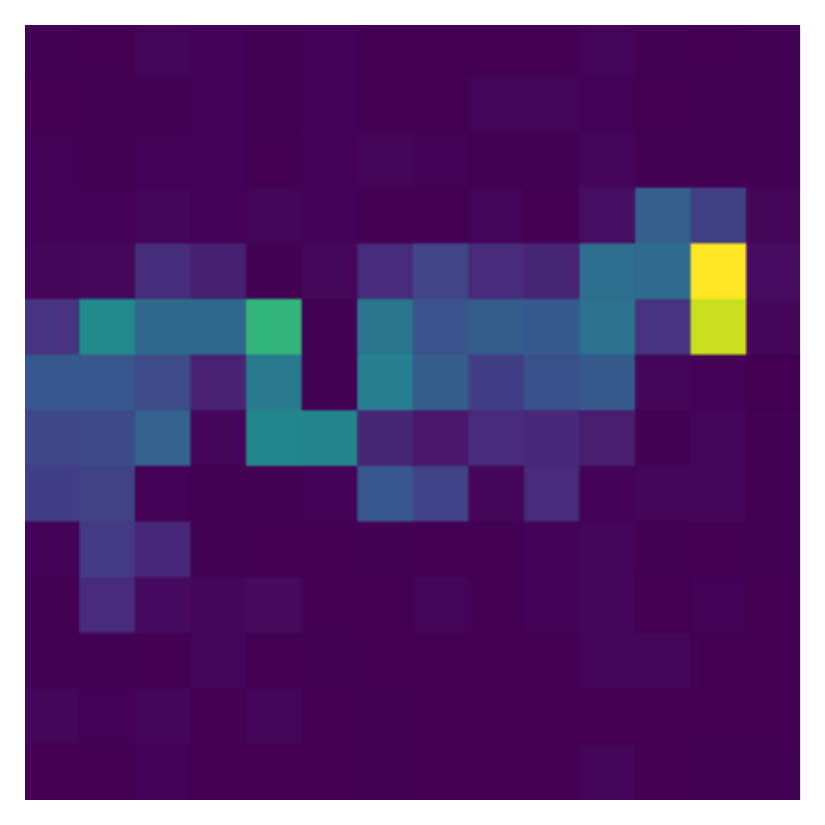} 
          \myfigc{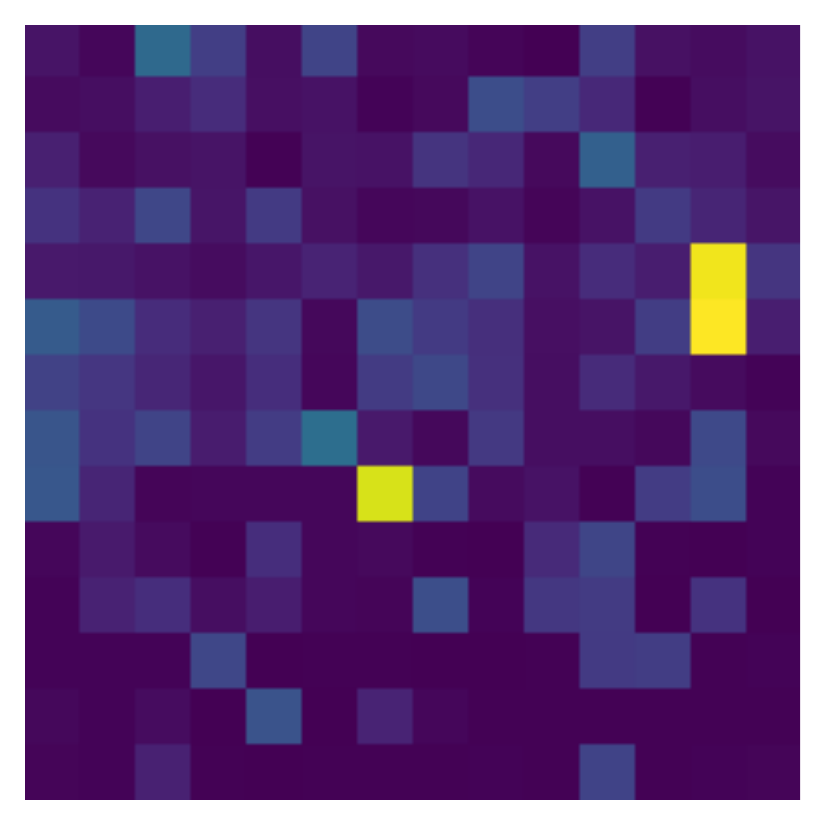} 
          \myfigc{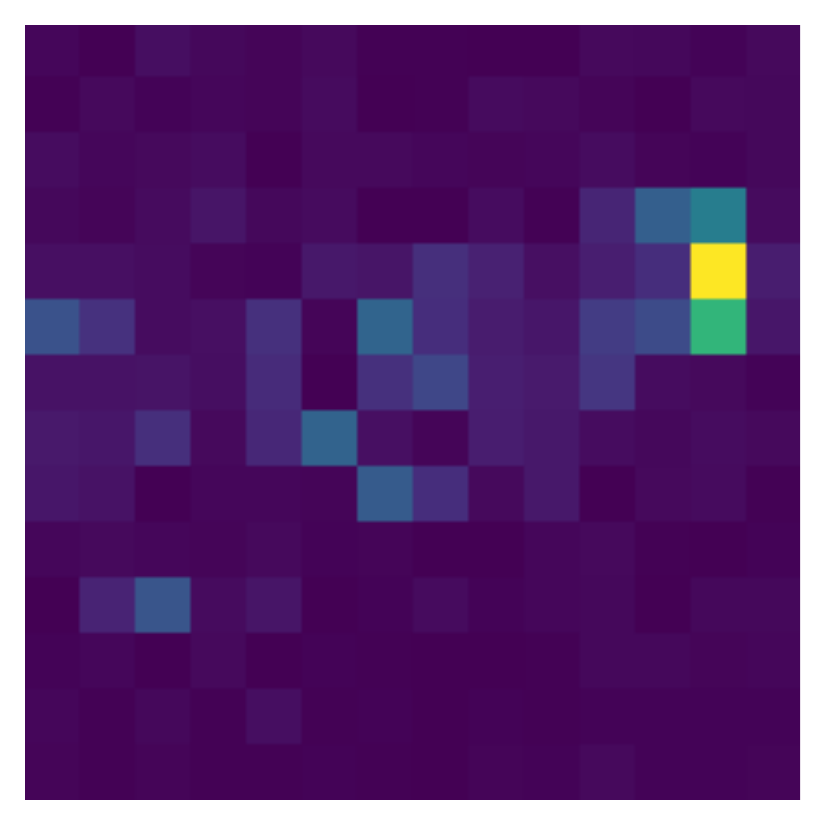} 
          \myfigc{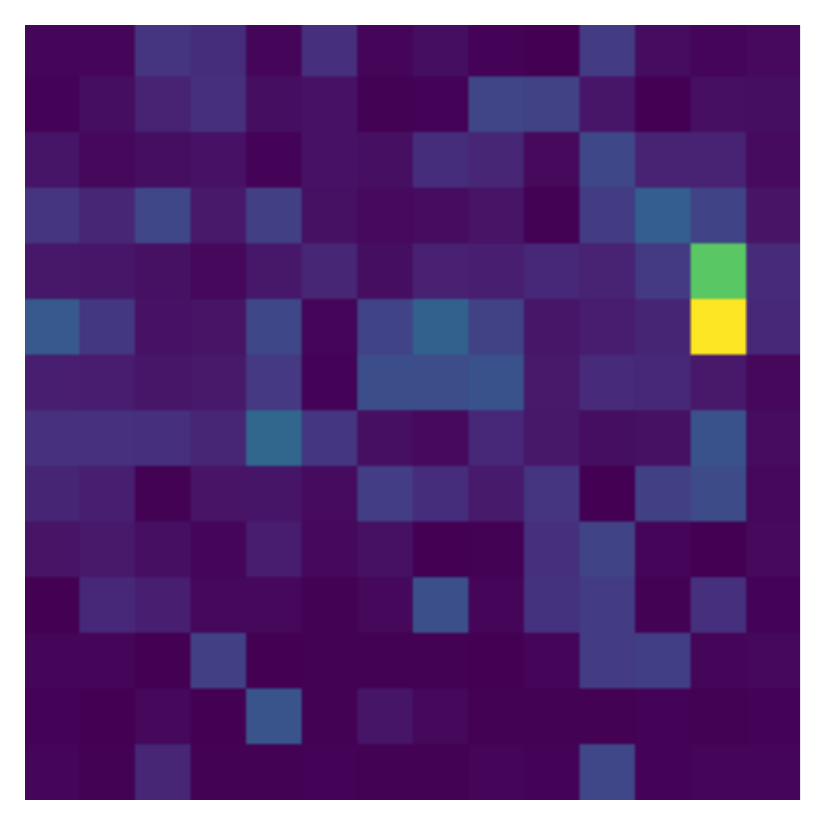} \\[-1em]
          
          ~\hspace{0.19\linewidth} \hfill
          \myfigc{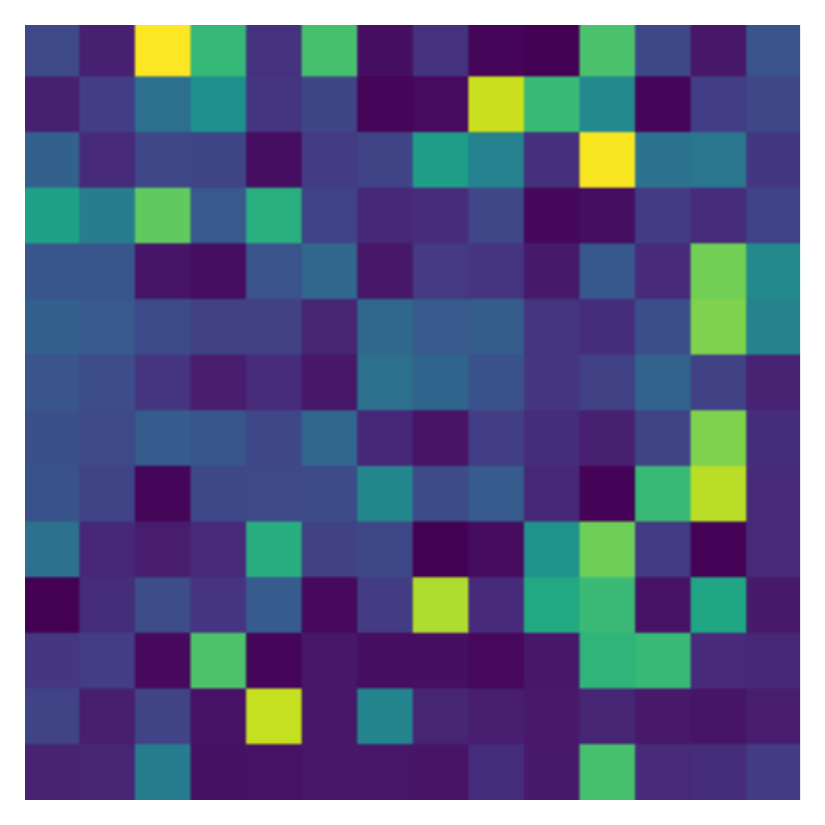} 
          \myfigc{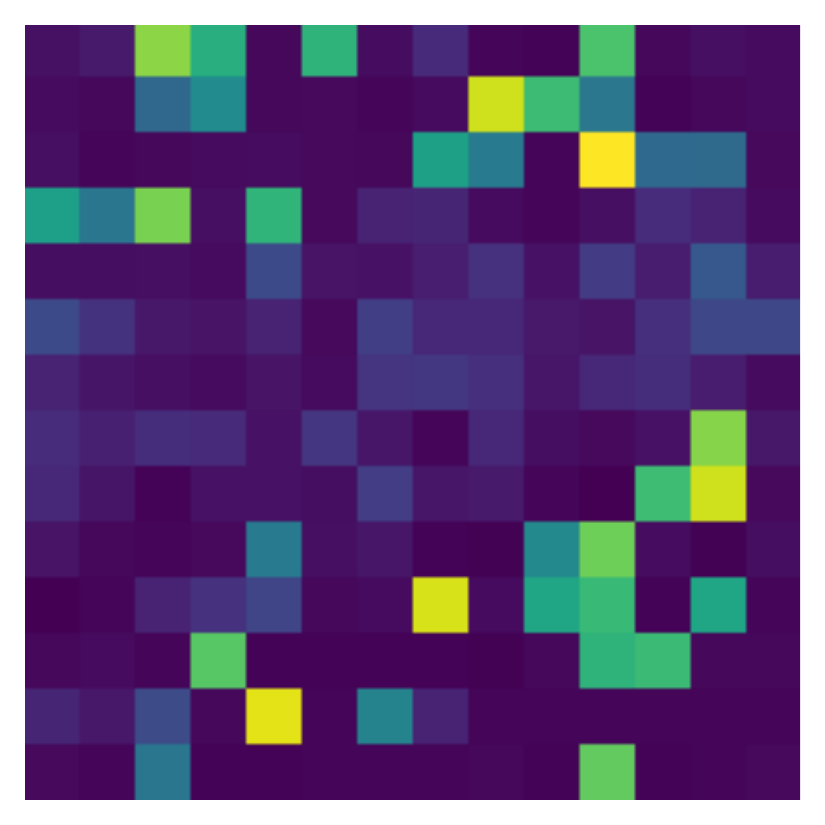} 
          \myfigc{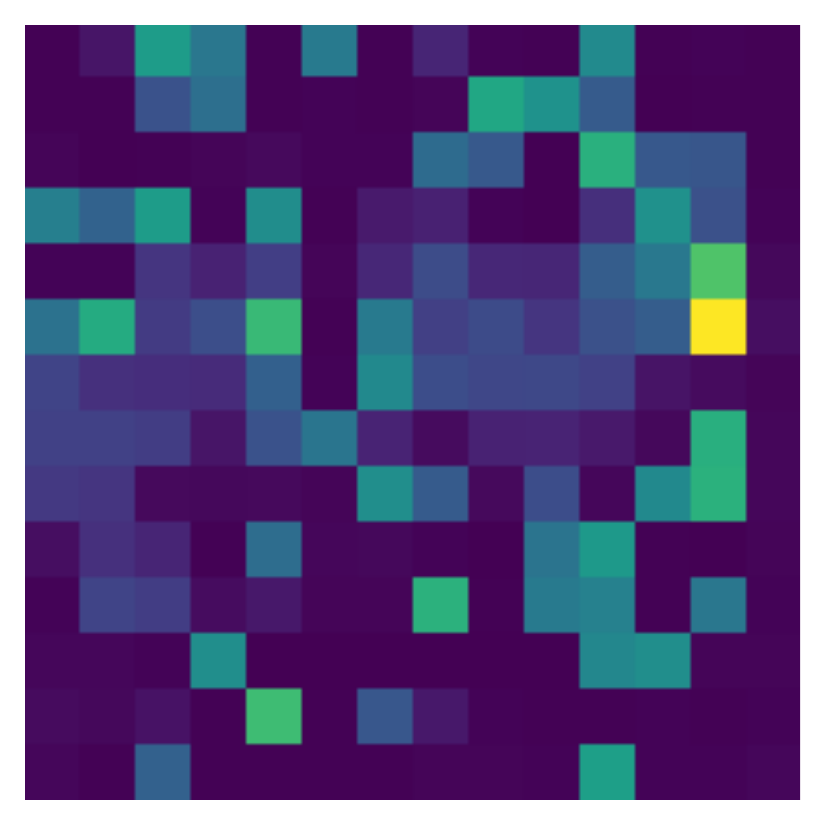} 
          \myfigc{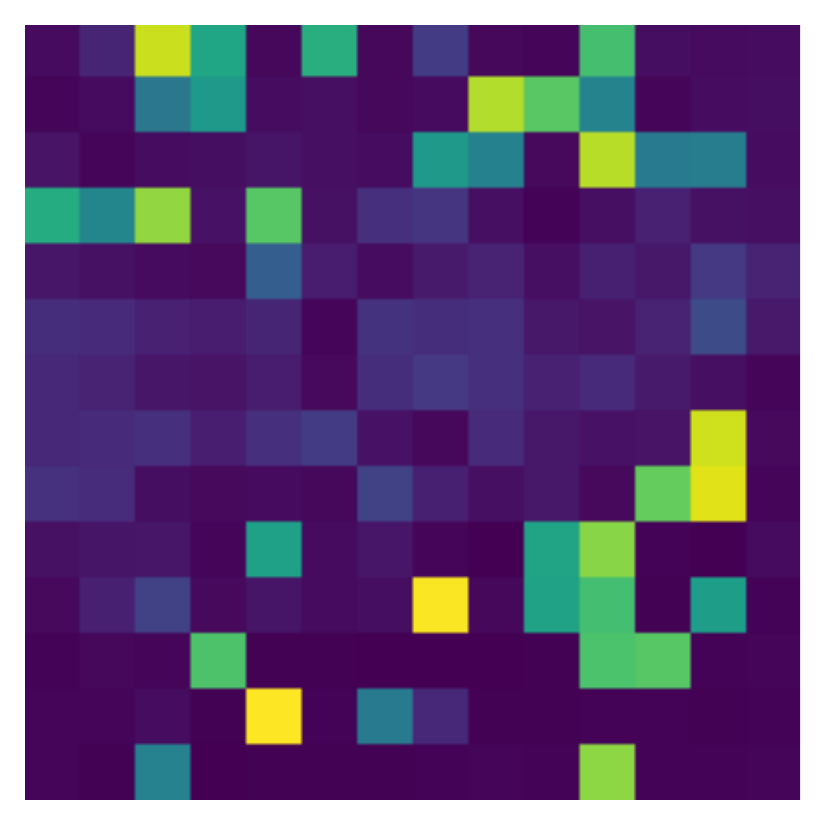} \\[-1em]
          
          \myfigc{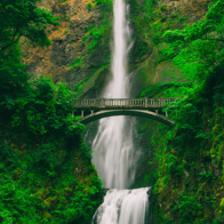} \hfill
          \myfigc{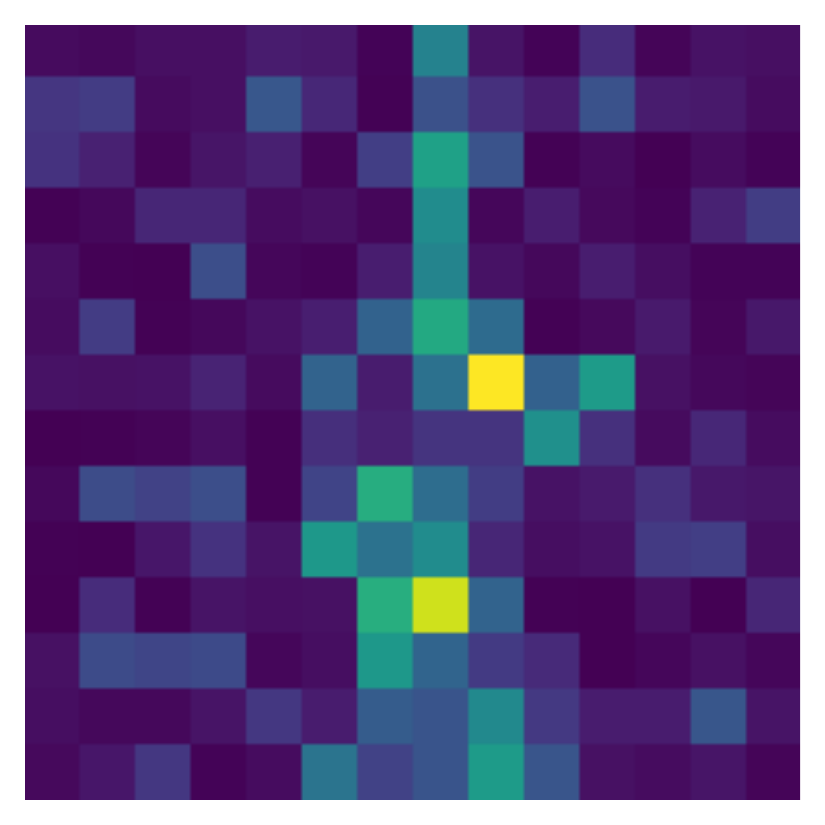} 
          \myfigc{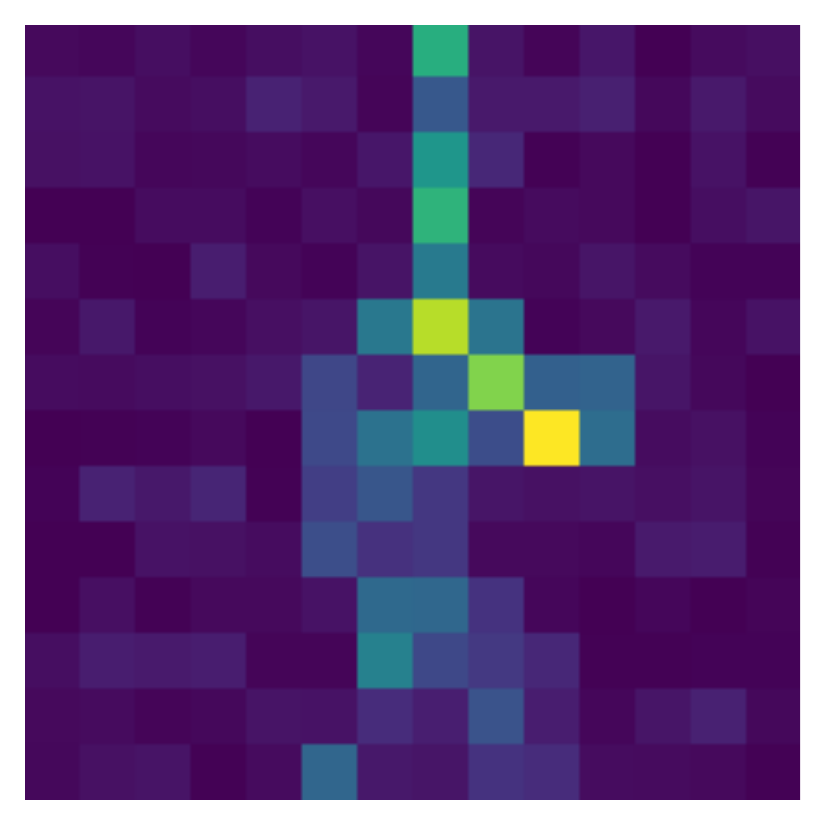} 
          \myfigc{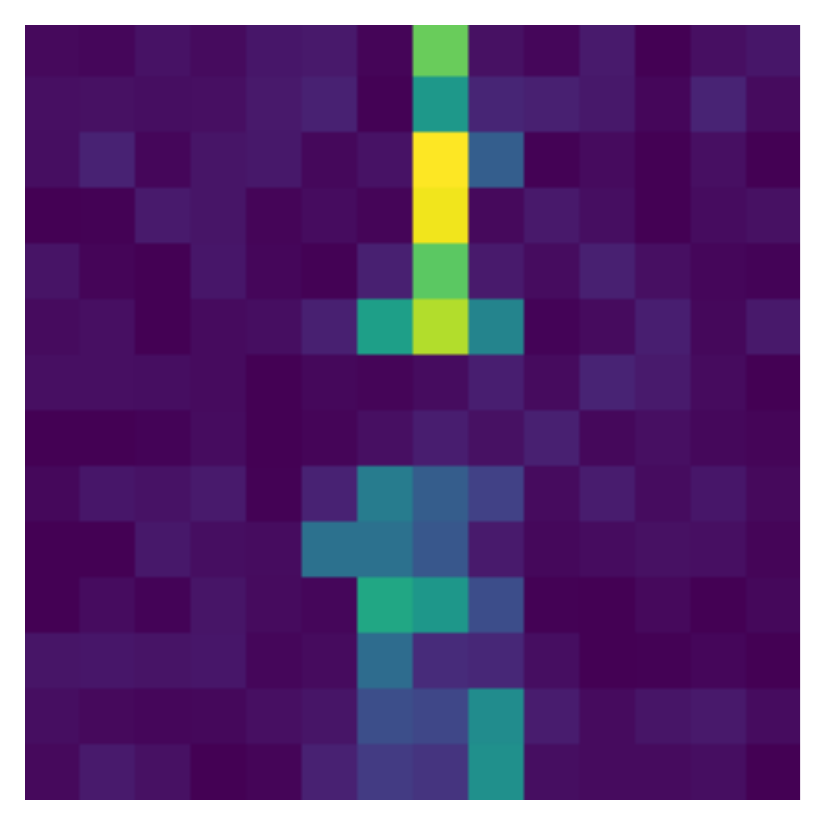} 
          \myfigc{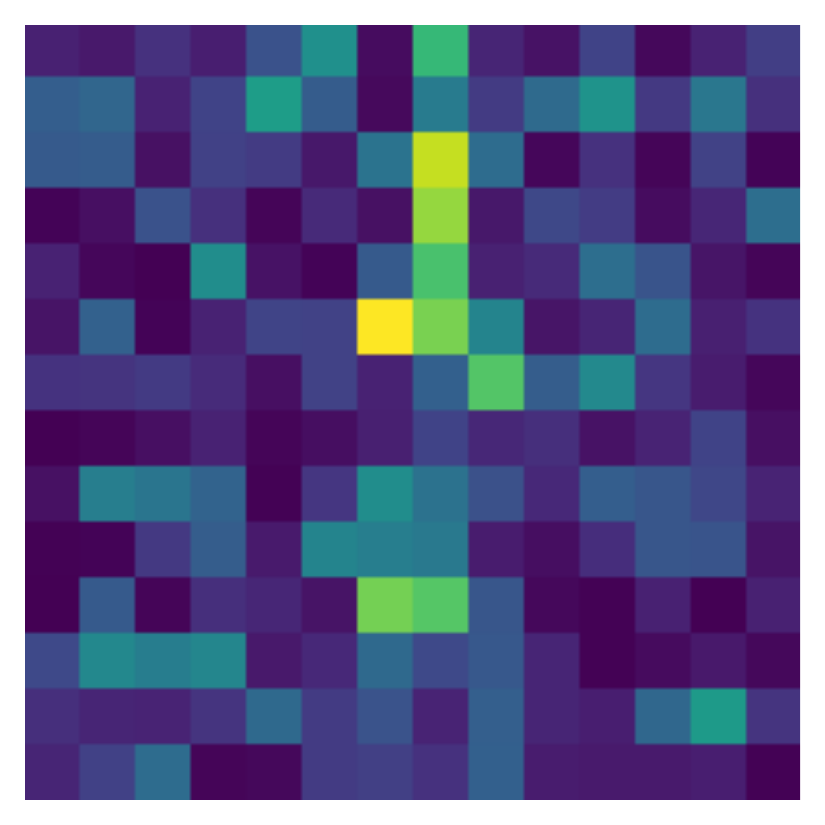} \\[-1em]
          
          ~\hspace{0.19\linewidth} \hfill
          \myfigc{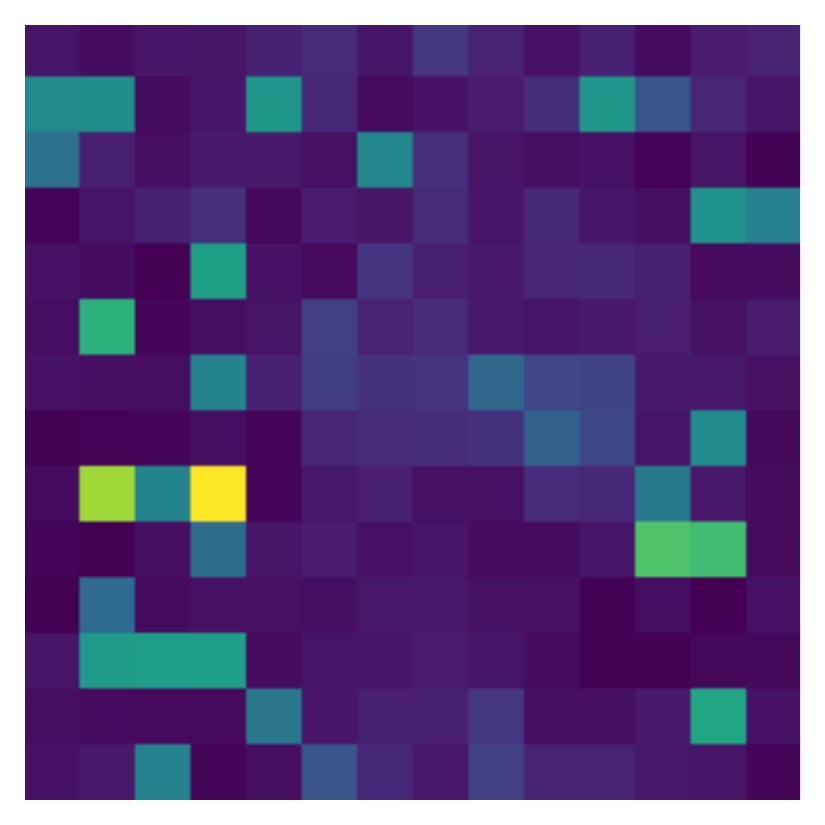} 
          \myfigc{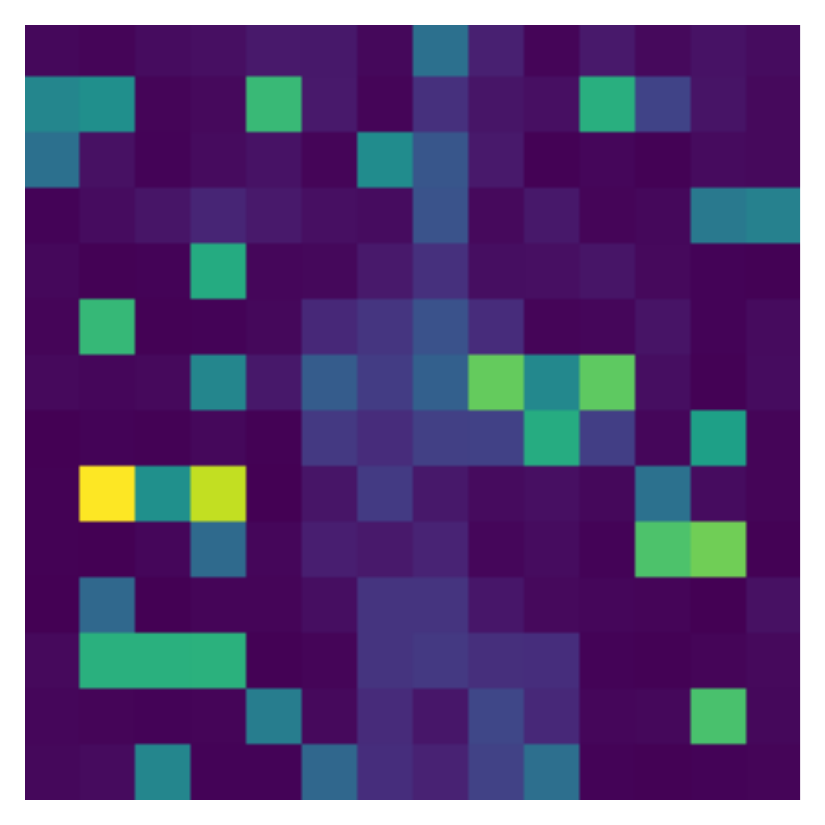} 
          \myfigc{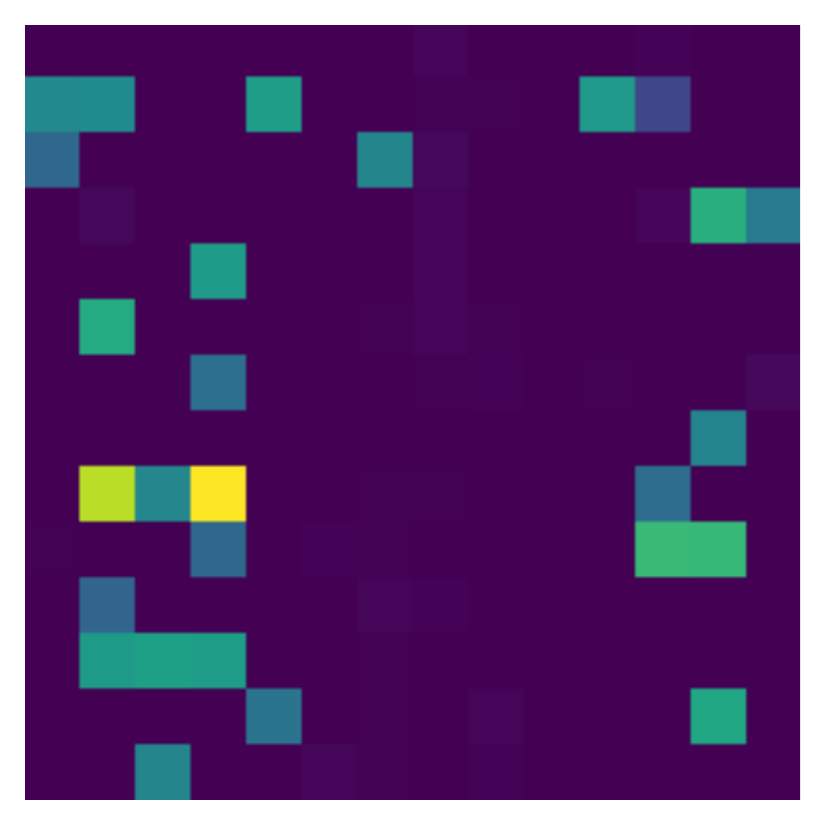} 
          \myfigc{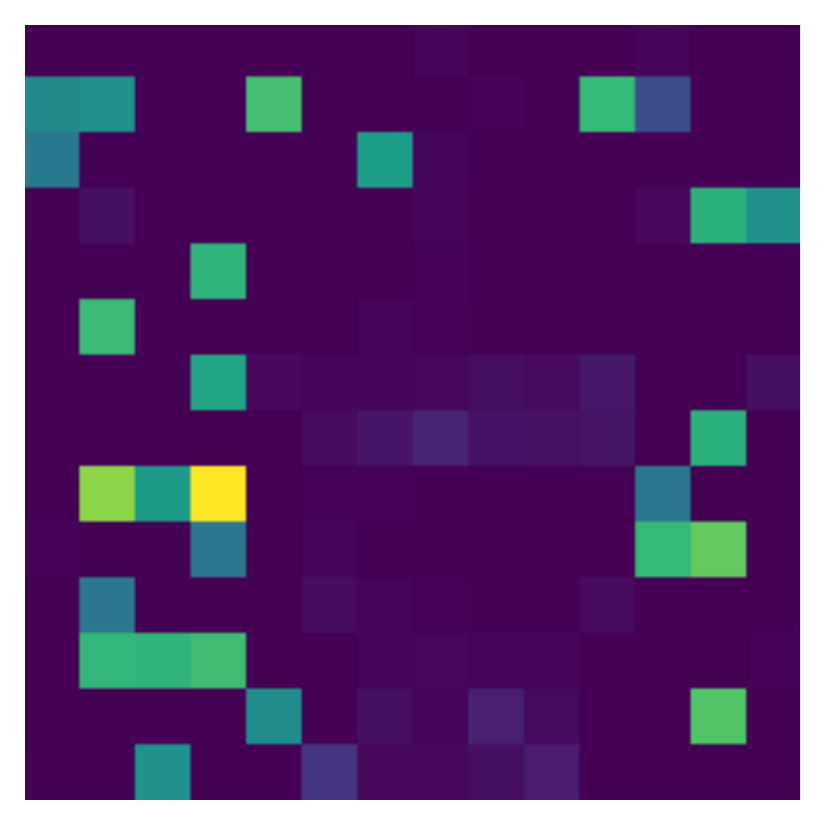} \\[-1em]
\end{minipage}}
    \caption{Visualization of the attention maps in the class-attention stage, obtained with a XXS model. For each image we present two rows: the top row correspond to the four heads of the attention maps associated with the first CA layer. The bottom row correspond to the four heads of the second CA layer. 
    \label{fig:visu_attention_appendix1}}
\end{figure}

\newcommand \inssal[2]  {
\centering \scalebox{0.7}{\scriptsize #1} \\[0pt]
\includegraphics[width=0.48\linewidth]{figs/images/images/#2} 
\includegraphics[width=0.48\linewidth]{figs/images/Attention/attention_mean_#2} \\[-7pt]
}

\begin{figure*}[p]
\vspace{-62pt}
\hspace{-55pt}
\scalebox{1.3}{
\begin{minipage}{0.325\linewidth}
\inssal{fountain (57\%)}{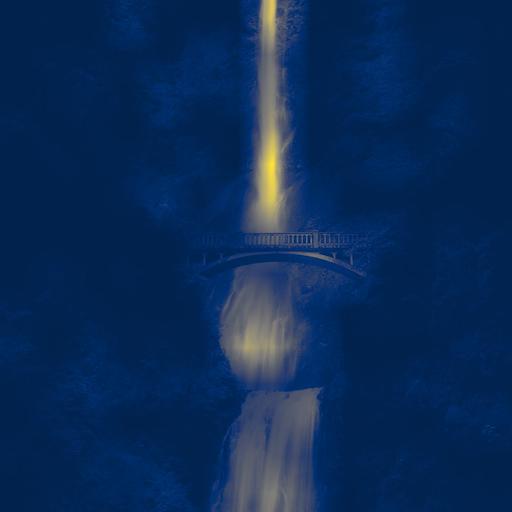}
\inssal{American alligator (77\%)}{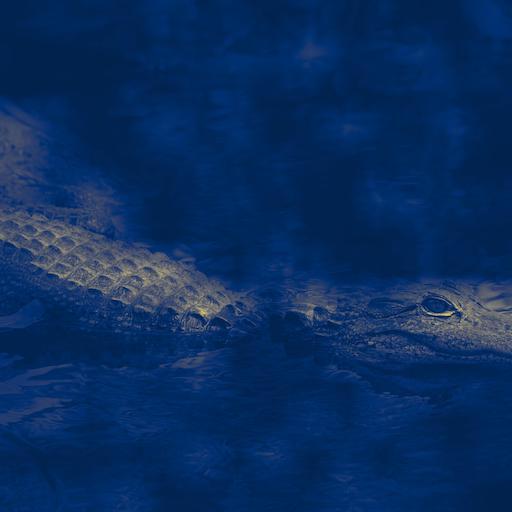}
\inssal{African chameleon (24\%), leaf beetle (12\%)}{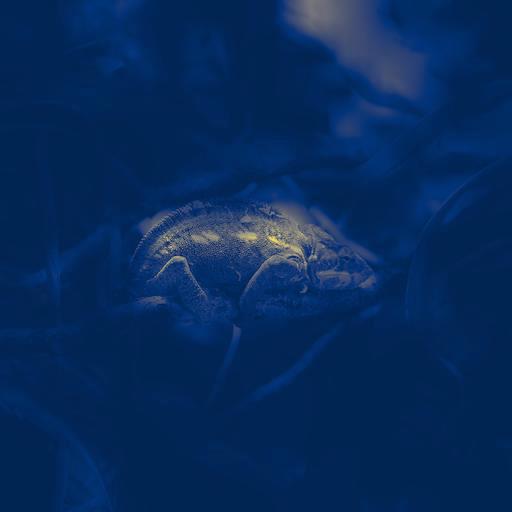}
\inssal{lakeside (40\%), alp (19\%),  valley(17\%)}{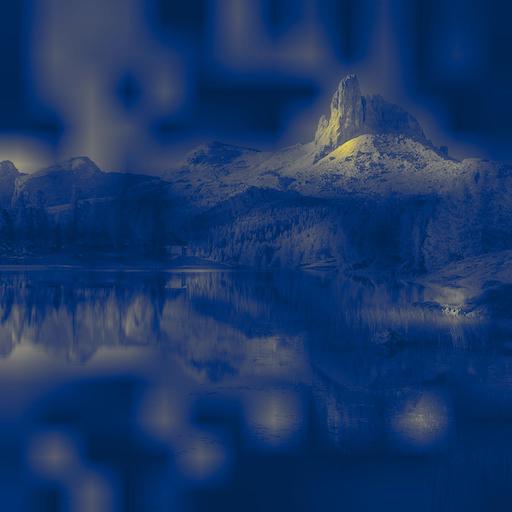}
\inssal{ambulance (24\%), traffic light (23\%)}{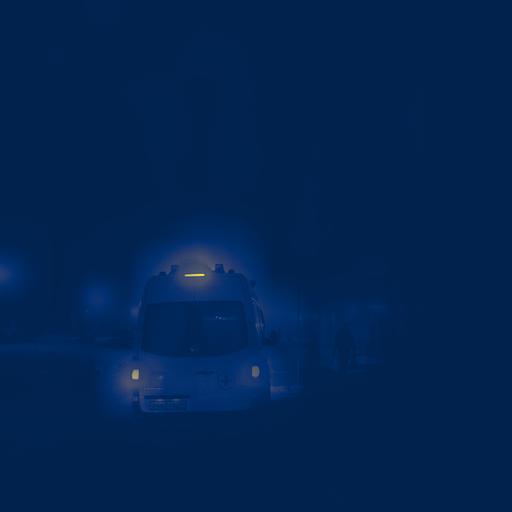}
\inssal{barbershop (84\%)}{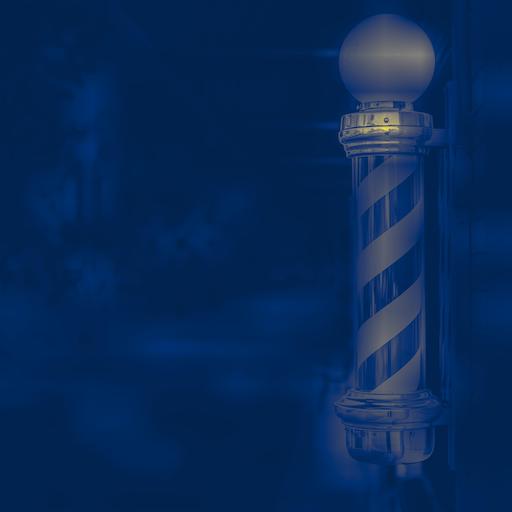}
\inssal{soap dispenser (40\%), lavabo (39\%)}{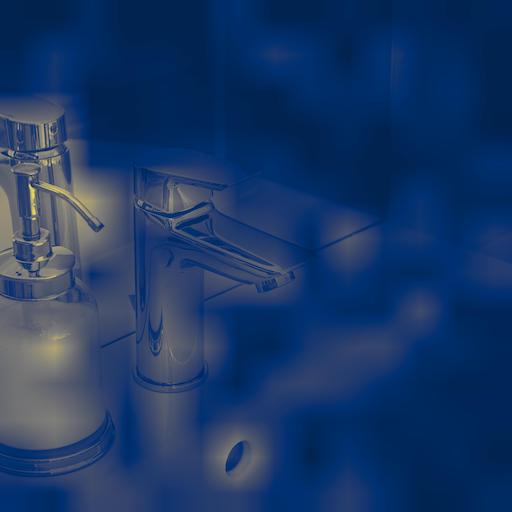}
\inssal{70 \% volcano}{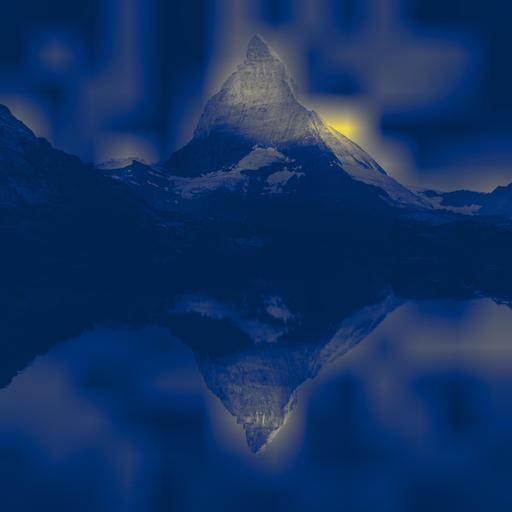}
\end{minipage}
\hfill
\begin{minipage}{0.325\linewidth}
\inssal{racket (73\%), tennis ball (10\%) }{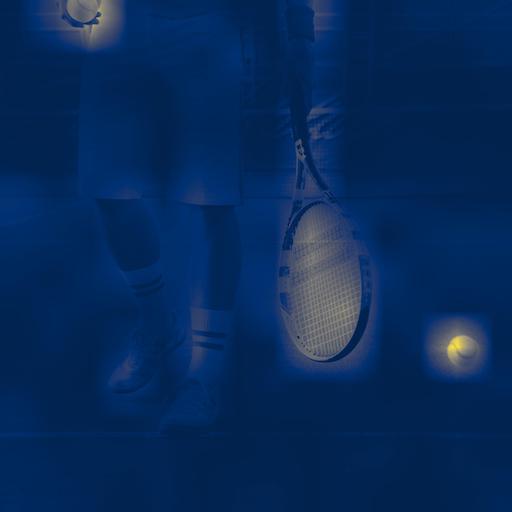}
\inssal{golf ball (71\%)}{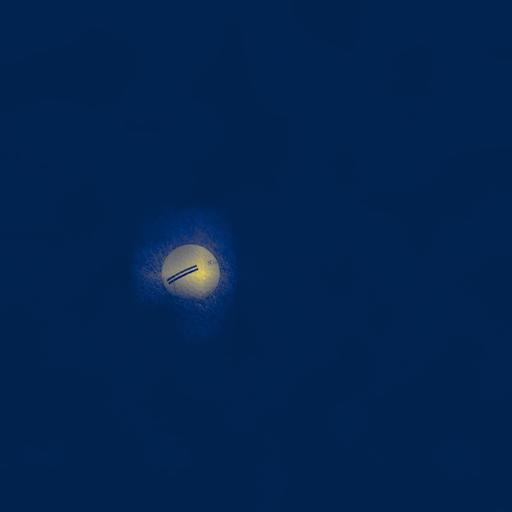}
\inssal{African elephant (57\%), water buffalo (15\%)}{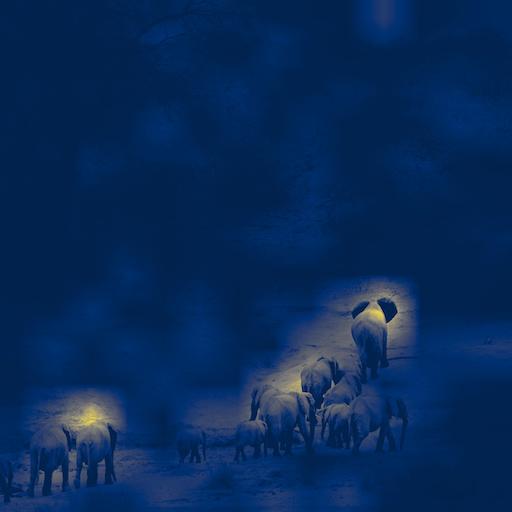}
\inssal{viaduct (87\%)}{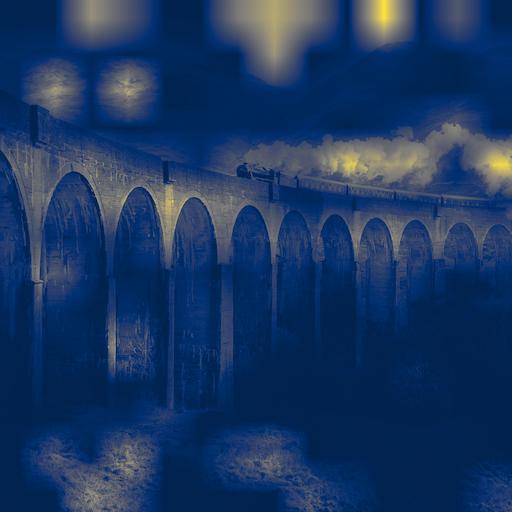}
\inssal{baboon (22\%), black bear (17\%), hyena (16\%)}{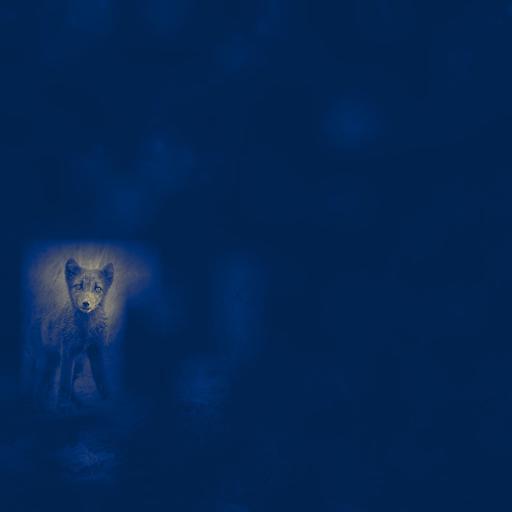}
\inssal{\makebox{convertible (27\%), taxi (15\%), sport car (12\%), wagon (11\%)}}{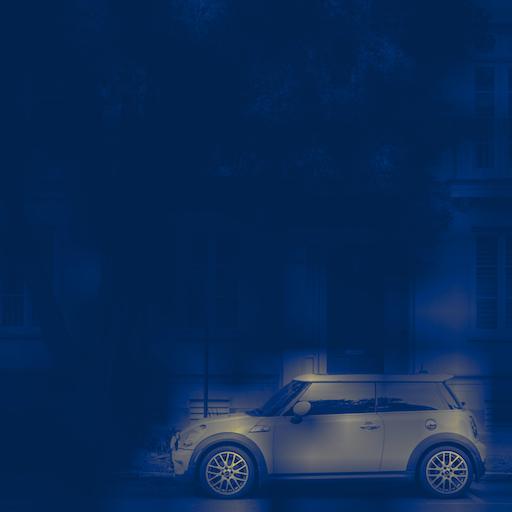}
\inssal{catamaran (61\%)}{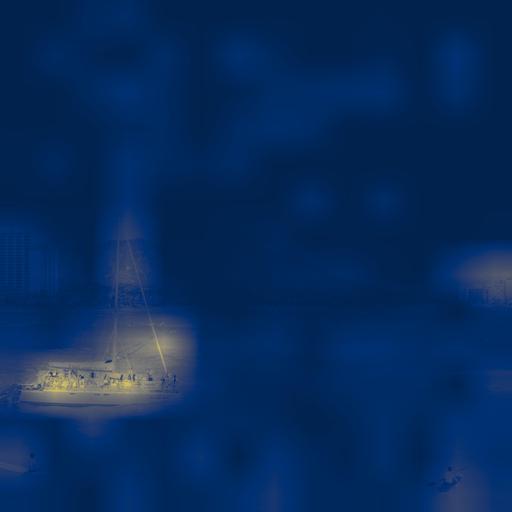}
\inssal{barrow (50\%), plow (26\%)}{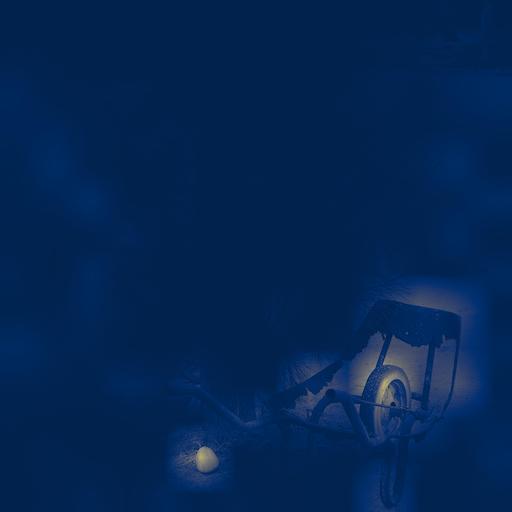}
\end{minipage}
\hfill
\begin{minipage}{0.325\linewidth}
\inssal{monarch butterfly (80\%)}{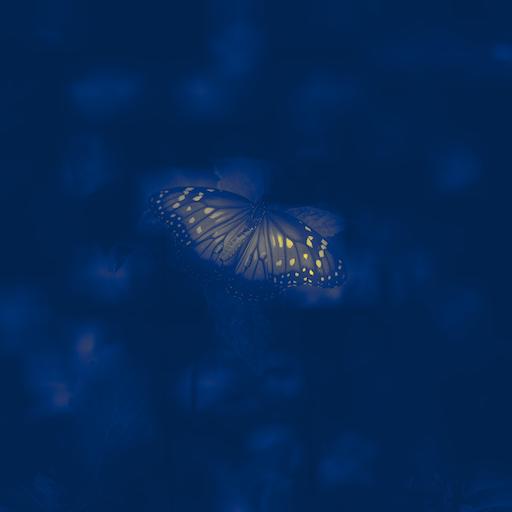}
\inssal{minibus (21\%), recreational vehicle (18\%)}{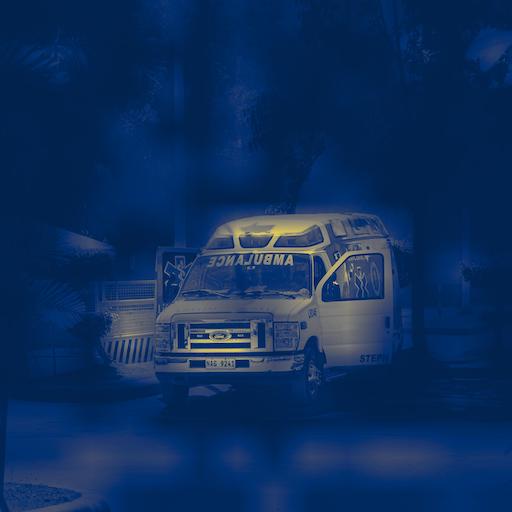}
\inssal{cup (43\%), notebook computer (19\%)}{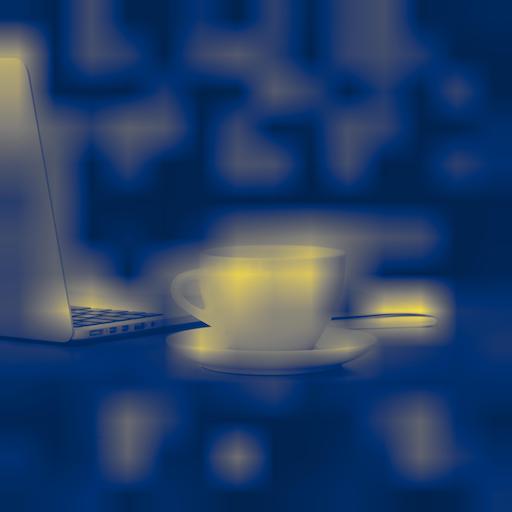}
\inssal{airliner (83\%)}{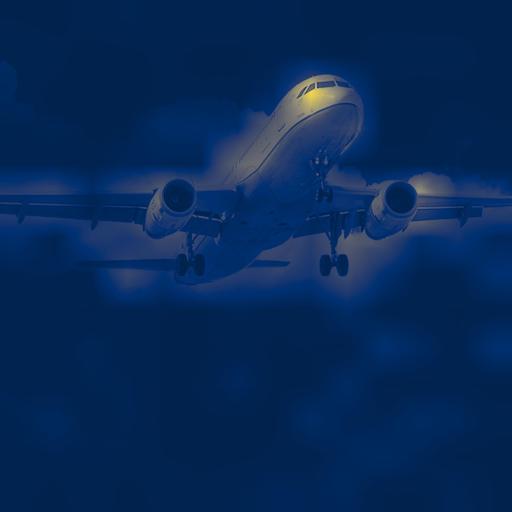}
\inssal{lakeside (24\%), coral fungus (16\%), coral reef (10\%)}{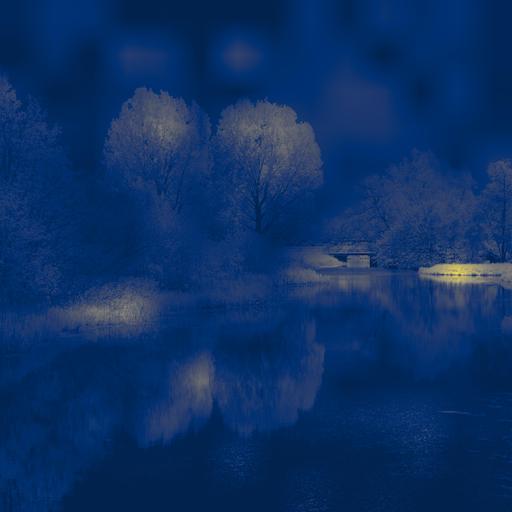}
\inssal{plate (50\%), carbonara (18\%)}{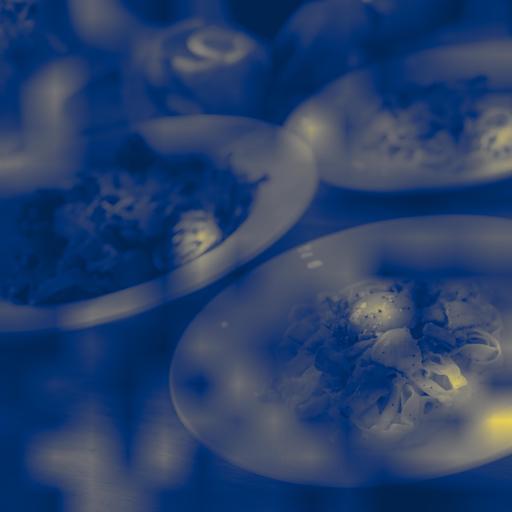}
\inssal{flagpole (14\%), dam (11\%)}{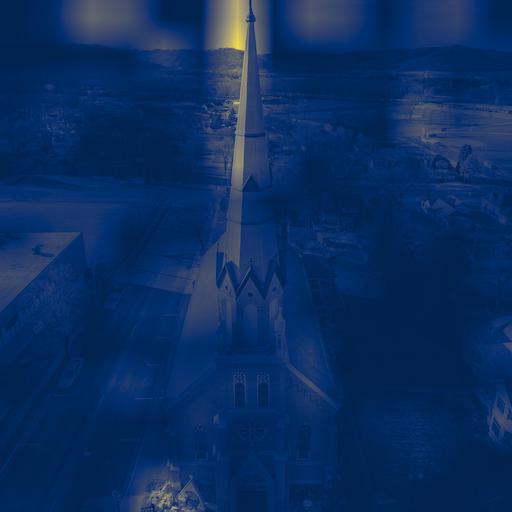}
\inssal{television (69\%)}{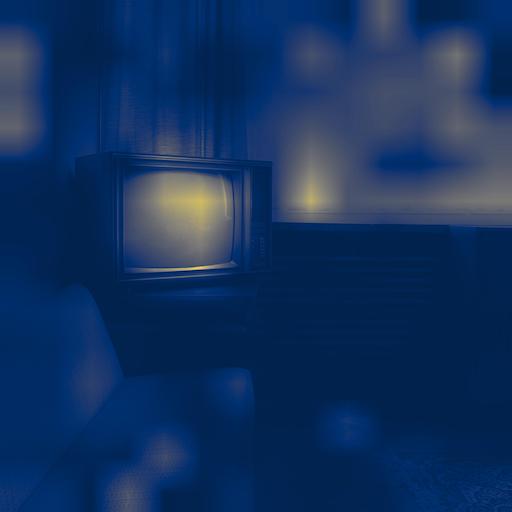}
\end{minipage}}
\caption{Illustration of the regions of focus of a \OURS-XXS model, according to the response of the first class-attention layer. 
\label{fig:saliency}}
\end{figure*}

\subsection{Attention map}

In Figure~\ref{fig:visu_attention_appendix1} we show the attention maps associated with the individual 4 heads of a XXS \OURS model, and for the two layers of class-attention. 
In \OURS and in contrast to ViT, the class-attention stage is the only one where there is some interaction between the class token and the patches, therefore it conveniently concentrates all the spatial-class relationship. 
We make two observations: 
\begin{itemize}
    \item The first class-attention layer clearly focuses on the object of interest, corresponding to the main part of the image on which the classification decision is performed (either correct or incorrect). 
    In this layer, the different heads focus either on the same or on complementary parts of the objects. This is especially visible for the waterfall image;
    \item The second class-attention layer seems to focus more on the context, or at least the image more globally. 
\end{itemize}

\subsection{Illustration of saliency in class-attention}

In figure~\ref{fig:saliency} we provide more vizualisations for a XXS model. They are just illustration of the saliency that one may extract from the first class-attention layer. As discussed previously this layer is the one that, empirically, is the most related to the object of interest. 
To produce these visual representations we simply average the attention maps from the different heads (depicted in Figure~\ref{fig:visu_attention_appendix1}), and upsample the resulting map to the image size. We then modulate the gray-level image with the strength of the attention after normalizing it with a simple rule of the form $(x-x_\mathrm{min})/(x_\mathrm{max}-x_\mathrm{min})$. 
We display the resulting image with \texttt{cividis} colormap.

For each image we show this saliency map and provides all the class for which the model assigns a probability higher than 10\%. These visualizations illustrate how the model can focus on two distinct regions (like racket and tennis ball on the top row/center). We can also observe some failure cases, like the top of the church classified as a flagpole. 

\section{Related work }
\label{sec:related} 

Since AlexNet \cite{Krizhevsky2012AlexNet}, convolutional neural networks (CNN) are the standard in image classification~\cite{He2016ResNet,tan2019efficientnet,Touvron2019FixRes}, and more generally in computer vision. 
While a deep CNN can theoretically model long range interaction between pixels across many layers, there has been research in increasing the range of interactions within a single layer. Some approaches adapt the receptive field of convolutions dynamically \cite{dai2017deformable,Li2019SelectiveKN}. 
At another end of the spectrum, attention can be viewed as a general form of non-local means, which was used in filtering (e.g.~denoising \cite{buades2005non}), and more recently in conjunction with convolutions \cite{Wang2018NonlocalNN}. 
Various other attention mechanism have been used successfully to give a global view in conjunction with (local) convolutions \cite{bello2019attention,Ramachandran2019StandAloneSI,chen2020dynamic,Zhang2020ResNeStSN,zhao2020exploring}, most mimic \textit{squeeze-and-excitate} \cite{Hu2017SENet} for leveraging global features. 
Lastly, LambdaNetworks \cite{bello2021lambdanetworks} decomposes attention into an approximated content attention and a batch-amortized positional attention component. 

Hybrid architectures combining CNNs and transformers blocks have also been used on ImageNet~\cite{Srinivas2021BottleneckTF,wu2020visual} and on COCO~\cite{carion2020end}. Originally, transformers without convolutions were applied on pixels directly~\cite{parmar2018image}, even scaling to hundred of layers~\cite{child2019generating}, but did not perform at CNNs levels. %
More recently, a transformer architecture working directly on small patches has obtained state of the art results on ImageNet~\cite{dosovitskiy2020image}.
Nevertheless, the state of the art has since returned to CNNs~\cite{Brock2021HighPerformanceLI,Pham2020MetaPL}. 
While some small improvements have been applied on the transformer architecture with encouraging results \cite{Yuan2021TokenstoTokenVT}, their performance is below the one of  DeiT~\cite{Touvron2020TrainingDI},  which uses a vanilla ViT architecture.

\paragraph{Encoder/decoder architectures.} Transformers were originally introduced for machine translation \cite{vaswani2017attention} with encoder-decoder models, and gained popularity as masked language model encoders (BERT) \cite{devlin2018bert,liu2019roberta}. % TODO cite ELECTRA + T5?
They yielded impressive results as scaled up language models, e.g. GPT-2 and 3 \cite{radford2019language,brown2020language}. 
They became a staple in speech recognition too \cite{luscher2019transformers,karita2019comparative}, being it in encoder and sequence criterion or encoder-decoder seq2seq \cite{synnaeve2019end} conformations, and hold the state of the art to this day \cite{xu2020self,zhang2020pushing} with models 36 blocks deep. 
Note, transforming only the class token with frozen trunk embeddings in \OURS is reminiscent of non-autoregressive encoder-decoders \cite{gu2017non,lee2018deterministic}, where a whole sequence (we have only one prediction) is produced at once by iterative refinements.

\paragraph{Deeper architectures}
 usually lead to better performance ~\cite{He2016ResNet,Simonyan2015VGG,Szegedy2015Goingdeeperwithconvolutions}, however this complicates their training process~\cite{Srivastava2015HighwayN,Srivastava2015TrainingVD}. One must adapt the architecture and the optimization procedure to train them correctly.
Some approaches focus on the initialization schemes~\cite{Glorot2010UnderstandingTD,He2016ResNet,Xiao2018DynamicalIA}, others on multiple stages training~\cite{Romero2015FitNetsHF,Simonyan2015VGG}, multiple loss at different depth~\cite{Szegedy2015Goingdeeperwithconvolutions}, adding components in the architecture~\cite{Bachlechner2020ReZeroIA,Zhang2019FixupIR} or regularization~\cite{Huang2016DeepNW}.
As pointed in our paper, in that respect our LayerScale approach is more related to Rezero~\cite{Bachlechner2020ReZeroIA} and  Skipinit~\cite{de2020batch}, Fixup \cite{Zhang2019FixupIR}, and T-Fixup \cite{huang2020improving}.

\section{Conclusion}
\label{sec:conclusion}

In this paper, we have shown how train deeper transformer-based image classification neural networks when training on Imagenet only. 
We have also introduced the simple yet effective \OURS architecture designed in the spirit of encoder/decoder architectures. 
Our work further demonstrates that transformer models offer a competitive alternative to the best convolutional neural networks when considering trade-offs between accuracy and  complexity. 

\section{Acknowledgments}

Thanks to Jakob Verbeek for his detailled feedback on an earlier version of this paper, to Alaa El-Nouby for fruitful discussions, to Mathilde Caron for suggestions regarding the vizualizations, and to Ross Wightman for the Timm library and the insights that he shares with the community.

\begingroup
    \small
    \bibliographystyle{ieee_fullname}
    \bibliography{egbib}

\begin{thebibliography}{10}\itemsep=-1pt

\bibitem{ba2016layer}
Jimmy~Lei Ba, Jamie~Ryan Kiros, and Geoffrey~E Hinton.
\newblock Layer normalization.
\newblock {\em arXiv preprint arXiv:1607.06450}, 2016.

\bibitem{Bachlechner2020ReZeroIA}
Thomas~C. Bachlechner, Bodhisattwa~Prasad Majumder, H.~H. Mao, G. Cottrell, and
  Julian McAuley.
\newblock Rezero is all you need: Fast convergence at large depth.
\newblock {\em arXiv preprint arXiv:2003.04887}, 2020.

\bibitem{bello2021lambdanetworks}
Irwan Bello.
\newblock Lambdanetworks: Modeling long-range interactions without attention.
\newblock In {\em International Conference on Learning Representations}, 2021.

\bibitem{bello2019attention}
Irwan Bello, Barret Zoph, Ashish Vaswani, Jonathon Shlens, and Quoc~V Le.
\newblock Attention augmented convolutional networks.
\newblock In {\em Conference on Computer Vision and Pattern Recognition}, 2019.

\bibitem{berman2019multigrain}
Maxim Berman, Herv{\'{e}} J{\'{e}}gou, Andrea Vedaldi, Iasonas Kokkinos, and
  Matthijs Douze.
\newblock Multigrain: a unified image embedding for classes and instances.
\newblock {\em arXiv preprint arXiv:1902.05509}, 2019.

\bibitem{Beyer2020ImageNetReal}
Lucas Beyer, Olivier~J. H{\'e}naff, Alexander Kolesnikov, Xiaohua Zhai, and
  Aaron van~den Oord.
\newblock Are we done with imagenet?
\newblock {\em arXiv preprint arXiv:2006.07159}, 2020.

\bibitem{brock2021characterizing}
Andrew Brock, Soham De, and Samuel~L Smith.
\newblock Characterizing signal propagation to close the performance gap in
  unnormalized resnets.
\newblock {\em arXiv preprint arXiv:2101.08692}, 2021.

\bibitem{Brock2021HighPerformanceLI}
A. Brock, Soham De, S.~L. Smith, and K. Simonyan.
\newblock High-performance large-scale image recognition without normalization.
\newblock {\em arXiv preprint arXiv:2102.06171}, 2021.

\bibitem{brown2020language}
Tom~B Brown, Benjamin Mann, Nick Ryder, Melanie Subbiah, Jared Kaplan, Prafulla
  Dhariwal, Arvind Neelakantan, Pranav Shyam, Girish Sastry, Amanda Askell,
  et~al.
\newblock Language models are few-shot learners.
\newblock {\em arXiv preprint arXiv:2005.14165}, 2020.

\bibitem{buades2005non}
Antoni Buades, Bartomeu Coll, and J-M Morel.
\newblock A non-local algorithm for image denoising.
\newblock In {\em Conference on Computer Vision and Pattern Recognition}, 2005.

\bibitem{carion2020end}
Nicolas Carion, Francisco Massa, Gabriel Synnaeve, Nicolas Usunier, Alexander
  Kirillov, and Sergey Zagoruyko.
\newblock End-to-end object detection with transformers.
\newblock In {\em European Conference on Computer Vision}, 2020.

\bibitem{chen2020dynamic}
Yinpeng Chen, Xiyang Dai, Mengchen Liu, Dongdong Chen, Lu Yuan, and Zicheng
  Liu.
\newblock Dynamic convolution: Attention over convolution kernels.
\newblock In {\em Conference on Computer Vision and Pattern Recognition}, 2020.

\bibitem{child2019generating}
Rewon Child, Scott Gray, Alec Radford, and Ilya Sutskever.
\newblock Generating long sequences with sparse transformers.
\newblock {\em arXiv preprint arXiv:1904.10509}, 2019.

\bibitem{Cubuk2019RandAugmentPA}
Ekin~D. Cubuk, Barret Zoph, Jonathon Shlens, and Quoc~V. Le.
\newblock Randaugment: Practical automated data augmentation with a reduced
  search space.
\newblock {\em arXiv preprint arXiv:1909.13719}, 2019.

\bibitem{dai2017deformable}
Jifeng Dai, Haozhi Qi, Yuwen Xiong, Yi Li, Guodong Zhang, Han Hu, and Yichen
  Wei.
\newblock Deformable convolutional networks.
\newblock In {\em Conference on Computer Vision and Pattern Recognition}, 2017.

\bibitem{de2020batch}
Soham De and Samuel~L Smith.
\newblock Batch normalization biases residual blocks towards the identity
  function in deep networks.
\newblock {\em arXiv e-prints}, pages arXiv--2002, 2020.

\bibitem{deng2009imagenet}
Jia Deng, Wei Dong, Richard Socher, Li-Jia Li, Kai Li, and Li Fei-Fei.
\newblock Imagenet: A large-scale hierarchical image database.
\newblock In {\em Conference on Computer Vision and Pattern Recognition}, pages
  248--255, 2009.

\bibitem{devlin2018bert}
Jacob Devlin, Ming-Wei Chang, Kenton Lee, and Kristina Toutanova.
\newblock Bert: Pre-training of deep bidirectional transformers for language
  understanding.
\newblock {\em arXiv preprint arXiv:1810.04805}, 2018.

\bibitem{dosovitskiy2020image}
Alexey Dosovitskiy, Lucas Beyer, Alexander Kolesnikov, Dirk Weissenborn,
  Xiaohua Zhai, Thomas Unterthiner, Mostafa Dehghani, Matthias Minderer, Georg
  Heigold, Sylvain Gelly, et~al.
\newblock An image is worth 16x16 words: Transformers for image recognition at
  scale.
\newblock {\em International Conference on Learning Representations}, 2021.

\bibitem{el2021training}
Alaaeldin El-Nouby, Natalia Neverova, Ivan Laptev, and Herv{\'e} J{\'e}gou.
\newblock Training vision transformers for image retrieval.
\newblock {\em arXiv preprint arXiv:2102.05644}, 2021.

\bibitem{fan2019reducing}
Angela Fan, Edouard Grave, and Armand Joulin.
\newblock Reducing transformer depth on demand with structured dropout.
\newblock {\em arXiv preprint arXiv:1909.11556}, 2019.
\newblock ICLR 2020.

\bibitem{fan2020training}
Angela Fan, Pierre Stock, Benjamin Graham, Edouard Grave, R{\'e}mi Gribonval,
  Herv{\'e} J{\'e}gou, and Armand Joulin.
\newblock Training with quantization noise for extreme model compression.
\newblock {\em arXiv preprint arXiv:2004.07320}, 2020.

\bibitem{frankle2018lottery}
Jonathan Frankle and Michael Carbin.
\newblock The lottery ticket hypothesis: Finding sparse, trainable neural
  networks.
\newblock {\em arXiv preprint arXiv:1803.03635}, 2018.

\bibitem{Glorot2010UnderstandingTD}
Xavier Glorot and Yoshua Bengio.
\newblock Understanding the difficulty of training deep feedforward neural
  networks.
\newblock In {\em AISTATS}, 2010.

\bibitem{gu2017non}
Jiatao Gu, James Bradbury, Caiming Xiong, Victor~OK Li, and Richard Socher.
\newblock Non-autoregressive neural machine translation.
\newblock {\em arXiv preprint arXiv:1711.02281}, 2017.

\bibitem{Han2021TransformerIT}
Kai Han, An Xiao, Enhua Wu, Jianyuan Guo, Chunjing Xu, and Yunhe Wang.
\newblock Transformer in transformer.
\newblock {\em arXiv preprint arXiv:2103.00112}, 2021.

\bibitem{He2016ResNet}
Kaiming He, Xiangyu Zhang, Shaoqing Ren, and Jian Sun.
\newblock Deep residual learning for image recognition.
\newblock In {\em Conference on Computer Vision and Pattern Recognition}, June
  2016.

\bibitem{He2016IdentityMappings}
Kaiming He, Xiangyu Zhang, Shaoqing Ren, and Jian Sun.
\newblock Identity mappings in deep residual networks.
\newblock {\em arXiv preprint arXiv:1603.05027}, 2016.

\bibitem{hoffer2020augment}
Elad Hoffer, Tal Ben-Nun, Itay Hubara, Niv Giladi, Torsten Hoefler, and Daniel
  Soudry.
\newblock Augment your batch: Improving generalization through instance
  repetition.
\newblock In {\em Conference on Computer Vision and Pattern Recognition}, 2020.

\bibitem{Horn2018INaturalist}
Grant~Van Horn, Oisin {Mac Aodha}, Yang Song, Alexander Shepard, Hartwig Adam,
  Pietro Perona, and Serge~J. Belongie.
\newblock The inaturalist challenge 2018 dataset.
\newblock {\em arXiv preprint arXiv:1707.06642}, 2018.

\bibitem{Horn2019INaturalist}
Grant~Van Horn, Oisin {Mac Aodha}, Yang Song, Alexander Shepard, Hartwig Adam,
  Pietro Perona, and Serge~J. Belongie.
\newblock The inaturalist challenge 2019 dataset.
\newblock {\em arXiv preprint arXiv:1707.06642}, 2019.

\bibitem{Hu2017SENet}
Jie Hu, Li Shen, and Gang Sun.
\newblock Squeeze-and-excitation networks.
\newblock {\em arXiv preprint arXiv:1709.01507}, 2017.

\bibitem{Huang2016DeepNW}
Gao Huang, Yu Sun, Zhuang Liu, Daniel Sedra, and Kilian~Q. Weinberger.
\newblock Deep networks with stochastic depth.
\newblock In {\em European Conference on Computer Vision}, 2016.

\bibitem{huang2020improving}
Xiao~Shi Huang, Felipe Perez, Jimmy Ba, and Maksims Volkovs.
\newblock Improving transformer optimization through better initialization.
\newblock In {\em International Conference on Machine Learning}, pages
  4475--4483. PMLR, 2020.

\bibitem{ioffe15batchnorm}
Sergey Ioffe and Christian Szegedy.
\newblock Batch normalization: Accelerating deep network training by reducing
  internal covariate shift.
\newblock In {\em International Conference on Machine Learning}, 2015.

\bibitem{karita2019comparative}
Shigeki Karita, Nanxin Chen, Tomoki Hayashi, et~al.
\newblock A comparative study on transformer vs rnn in speech applications.
\newblock {\em arXiv preprint arXiv:1909.06317}, 2019.

\bibitem{kingma2018glow}
Diederik~P Kingma and Prafulla Dhariwal.
\newblock Glow: Generative flow with invertible 1x1 convolutions.
\newblock {\em arXiv preprint arXiv:1807.03039}, 2018.

\bibitem{Cars2013}
Jonathan Krause, Michael Stark, Jia Deng, and Li Fei-Fei.
\newblock 3d object representations for fine-grained categorization.
\newblock In {\em 4th International IEEE Workshop on 3D Representation and
  Recognition (3dRR-13)}, 2013.

\bibitem{Krizhevsky2009LearningML}
Alex Krizhevsky.
\newblock Learning multiple layers of features from tiny images.
\newblock Technical report, CIFAR, 2009.

\bibitem{Krizhevsky2012AlexNet}
Alex Krizhevsky, Ilya Sutskever, and Geoffrey~E. Hinton.
\newblock Imagenet classification with deep convolutional neural networks.
\newblock In {\em NIPS}, 2012.

\bibitem{lee2018deterministic}
Jason Lee, Elman Mansimov, and Kyunghyun Cho.
\newblock Deterministic non-autoregressive neural sequence modeling by
  iterative refinement.
\newblock {\em arXiv preprint arXiv:1802.06901}, 2018.

\bibitem{Li2019SelectiveKN}
Xiang Li, Wenhai Wang, Xiaolin Hu, and Jian Yang.
\newblock Selective kernel networks.
\newblock {\em Conference on Computer Vision and Pattern Recognition}, 2019.

\bibitem{liu2019roberta}
Yinhan Liu, Myle Ott, Naman Goyal, Jingfei Du, Mandar Joshi, Danqi Chen, Omer
  Levy, Mike Lewis, Luke Zettlemoyer, and Veselin Stoyanov.
\newblock Roberta: A robustly optimized bert pretraining approach.
\newblock {\em arXiv preprint arXiv:1907.11692}, 2019.

\bibitem{Loshchilov2017AdamW}
I. Loshchilov and F. Hutter.
\newblock Fixing weight decay regularization in adam.
\newblock {\em arXiv preprint arXiv:1711.05101}, 2017.

\bibitem{luscher2019transformers}
Christoph Lüscher, Eugen Beck, Kazuki Irie, et~al.
\newblock Rwth asr systems for librispeech: Hybrid vs attention.
\newblock {\em Interspeech 2019}, Sep 2019.

\bibitem{Nilsback08}
M-E. Nilsback and A. Zisserman.
\newblock Automated flower classification over a large number of classes.
\newblock In {\em Proceedings of the Indian Conference on Computer Vision,
  Graphics and Image Processing}, 2008.

\bibitem{parmar2018image}
Niki Parmar, Ashish Vaswani, Jakob Uszkoreit, Lukasz Kaiser, Noam Shazeer,
  Alexander Ku, and Dustin Tran.
\newblock Image transformer.
\newblock In {\em International Conference on Machine Learning}, pages
  4055--4064. PMLR, 2018.

\bibitem{Pham2020MetaPL}
H. Pham, Qizhe Xie, Zihang Dai, and Quoc~V. Le.
\newblock Meta pseudo labels.
\newblock {\em arXiv preprint arXiv:2003.10580}, 2020.

\bibitem{radford2019language}
Alec Radford, Jeffrey Wu, Rewon Child, David Luan, Dario Amodei, and Ilya
  Sutskever.
\newblock Language models are unsupervised multitask learners.

\bibitem{Radosavovic2020RegNet}
Ilija Radosavovic, Raj~Prateek Kosaraju, Ross~B. Girshick, Kaiming He, and
  Piotr Doll{\'a}r.
\newblock Designing network design spaces.
\newblock {\em Conference on Computer Vision and Pattern Recognition}, 2020.

\bibitem{Ramachandran2019StandAloneSI}
Prajit Ramachandran, Niki Parmar, Ashish Vaswani, I. Bello, Anselm Levskaya,
  and Jonathon Shlens.
\newblock Stand-alone self-attention in vision models.
\newblock In {\em Neurips}, 2019.

\bibitem{Recht2019ImageNetv2}
B. Recht, Rebecca Roelofs, L. Schmidt, and V. Shankar.
\newblock Do imagenet classifiers generalize to imagenet?
\newblock {\em arXiv preprint arXiv:1902.10811}, 2019.

\bibitem{Romero2015FitNetsHF}
A. Romero, Nicolas Ballas, S. Kahou, Antoine Chassang, C. Gatta, and Yoshua
  Bengio.
\newblock Fitnets: Hints for thin deep nets.
\newblock {\em arXiv preprint arXiv:1412.6550}, 2015.

\bibitem{Russakovsky2015ImageNet12}
Olga Russakovsky, Jia Deng, Hao Su, Jonathan Krause, Sanjeev Satheesh, Sean Ma,
  Zhiheng Huang, Andrej Karpathy, Aditya Khosla, Michael Bernstein,
  Alexander~C. Berg, and Li Fei-Fei.
\newblock Imagenet large scale visual recognition challenge.
\newblock {\em International journal of Computer Vision}, 2015.

\bibitem{Shazeer2020TalkingHeadsA}
Noam Shazeer, Zhenzhong Lan, Youlong Cheng, N. Ding, and L. Hou.
\newblock Talking-heads attention.
\newblock {\em arXiv preprint arXiv:2003.02436}, 2020.

\bibitem{Simonyan2015VGG}
K. Simonyan and A. Zisserman.
\newblock Very deep convolutional networks for large-scale image recognition.
\newblock In {\em International Conference on Learning Representations}, 2015.

\bibitem{Srinivas2021BottleneckTF}
A. Srinivas, Tsung-Yi Lin, Niki Parmar, Jonathon Shlens, P. Abbeel, and Ashish
  Vaswani.
\newblock Bottleneck transformers for visual recognition.
\newblock {\em arXiv preprint arXiv:2101.11605}, 2021.

\bibitem{Srivastava2015HighwayN}
R. Srivastava, Klaus Greff, and J. Schmidhuber.
\newblock Highway networks.
\newblock {\em arXiv preprint arXiv:1505.00387}, 2015.

\bibitem{Srivastava2015TrainingVD}
R. Srivastava, Klaus Greff, and J. Schmidhuber.
\newblock Training very deep networks.
\newblock In {\em NIPS}, 2015.

\bibitem{synnaeve2019end}
Gabriel Synnaeve, Qiantong Xu, Jacob Kahn, Tatiana Likhomanenko, Edouard Grave,
  Vineel Pratap, Anuroop Sriram, Vitaliy Liptchinsky, and Ronan Collobert.
\newblock End-to-end asr: from supervised to semi-supervised learning with
  modern architectures.
\newblock {\em arXiv preprint arXiv:1911.08460}, 2019.

\bibitem{Szegedy2015Goingdeeperwithconvolutions}
C. {Szegedy}, {Wei Liu}, {Yangqing Jia}, P. {Sermanet}, S. {Reed}, D.
  {Anguelov}, D. {Erhan}, V. {Vanhoucke}, and A. {Rabinovich}.
\newblock Going deeper with convolutions.
\newblock In {\em Conference on Computer Vision and Pattern Recognition}, 2015.

\bibitem{tan2019efficientnet}
Mingxing Tan and Quoc~V. Le.
\newblock Efficientnet: Rethinking model scaling for convolutional neural
  networks.
\newblock {\em arXiv preprint arXiv:1905.11946}, 2019.

\bibitem{Touvron2020TrainingDI}
Hugo Touvron, M. Cord, M. Douze, F. Massa, Alexandre Sablayrolles, and H.
  J{\'e}gou.
\newblock Training data-efficient image transformers \& distillation through
  attention.
\newblock {\em arXiv preprint arXiv:2012.12877}, 2020.

\bibitem{Touvron2019FixRes}
Hugo Touvron, Andrea Vedaldi, Matthijs Douze, and Herve Jegou.
\newblock Fixing the train-test resolution discrepancy.
\newblock {\em Neurips}, 2019.

\bibitem{Touvron2020FixingTT}
Hugo Touvron, Andrea Vedaldi, Matthijs Douze, and Herv{\'e} J{\'e}gou.
\newblock Fixing the train-test resolution discrepancy: Fixefficientnet.
\newblock {\em arXiv preprint arXiv:2003.08237}, 2020.

\bibitem{vaswani2017attention}
Ashish Vaswani, Noam Shazeer, Niki Parmar, Jakob Uszkoreit, Llion Jones,
  Aidan~N Gomez, Lukasz Kaiser, and Illia Polosukhin.
\newblock Attention is all you need.
\newblock {\em arXiv preprint arXiv:1706.03762}, 2017.

\bibitem{Wang2018NonlocalNN}
X. Wang, Ross~B. Girshick, A. Gupta, and Kaiming He.
\newblock Non-local neural networks.
\newblock {\em Conference on Computer Vision and Pattern Recognition}, 2018.

\bibitem{pytorchmodels}
Ross Wightman.
\newblock Pytorch image models.
\newblock \url{https://github.com/rwightman/pytorch-image-models}, 2019.

\bibitem{wu2020visual}
Bichen Wu, Chenfeng Xu, Xiaoliang Dai, Alvin Wan, Peizhao Zhang, Masayoshi
  Tomizuka, Kurt Keutzer, and Peter Vajda.
\newblock Visual transformers: Token-based image representation and processing
  for computer vision.
\newblock {\em arXiv preprint arXiv:2006.03677}, 2020.

\bibitem{Xiao2018DynamicalIA}
L. Xiao, Y. Bahri, Jascha Sohl-Dickstein, S. Schoenholz, and Jeffrey
  Pennington.
\newblock Dynamical isometry and a mean field theory of cnns: How to train 10,
  000-layer vanilla convolutional neural networks.
\newblock {\em arXiv preprint arXiv:1806.05393}, 2018.

\bibitem{Xie2020AdversarialEI}
Cihang Xie, Mingxing Tan, Boqing Gong, Jiang Wang, A. Yuille, and Quoc~V. Le.
\newblock Adversarial examples improve image recognition.
\newblock {\em Conference on Computer Vision and Pattern Recognition}, 2020.

\bibitem{Xie2017AggregatedRT}
Saining Xie, Ross~B. Girshick, Piotr Doll{\'a}r, Zhuowen Tu, and Kaiming He.
\newblock Aggregated residual transformations for deep neural networks.
\newblock {\em Conference on Computer Vision and Pattern Recognition}, 2017.

\bibitem{xu2020self}
Qiantong Xu, Alexei Baevski, Tatiana Likhomanenko, Paden Tomasello, Alexis
  Conneau, Ronan Collobert, Gabriel Synnaeve, and Michael Auli.
\newblock Self-training and pre-training are complementary for speech
  recognition.
\newblock {\em arXiv preprint arXiv:2010.11430}, 2020.

\bibitem{Yuan2021TokenstoTokenVT}
L. Yuan, Y. Chen, Tao Wang, Weihao Yu, Yujun Shi, F. Tay, Jiashi Feng, and S.
  Yan.
\newblock Tokens-to-token vit: Training vision transformers from scratch on
  imagenet.
\newblock {\em arXiv preprint arXiv:2101.11986}, 2021.

\bibitem{Zhang2019FixupIR}
Hongyi Zhang, Yann Dauphin, and Tengyu Ma.
\newblock Fixup initialization: Residual learning without normalization.
\newblock {\em arXiv preprint arXiv:1901.09321}, 2019.

\bibitem{Zhang2020ResNeStSN}
Hang Zhang, Chongruo Wu, Zhongyue Zhang, Yi Zhu, Zhi-Li Zhang, Haibin Lin, Yu e
  Sun, Tong He, Jonas Mueller, R. Manmatha, M. Li, and Alex Smola.
\newblock Resnest: Split-attention networks.
\newblock {\em arXiv preprint arXiv:2004.08955}, 2020.

\bibitem{zhang2020pushing}
Yu Zhang, James Qin, Daniel~S Park, Wei Han, Chung-Cheng Chiu, Ruoming Pang,
  Quoc~V Le, and Yonghui Wu.
\newblock Pushing the limits of semi-supervised learning for automatic speech
  recognition.
\newblock {\em arXiv preprint arXiv:2010.10504}, 2020.

\bibitem{zhao2020exploring}
Hengshuang Zhao, Jiaya Jia, and Vladlen Koltun.
\newblock Exploring self-attention for image recognition.
\newblock In {\em Conference on Computer Vision and Pattern Recognition}, 2020.

\end{thebibliography}
\endgroup

\clearpage

\appendix\newpage

\appendix

\clearpage
\counterwithin{figure}{section}
\counterwithin{table}{section}
\counterwithin{equation}{section}

\pagenumbering{Roman}  

\newpage
\vskip .375in
\begin{center}
{\Large \bf \inserttitle \\ \vspace{0.5cm} \large Appendix \par}
  \vspace*{24pt}
  {
  \par
  }
\end{center}

In this supplemental material, we provide variations on the architecture presented in our main paper. These experiments have guided some of our choices for the design of class-attention layers and LayerScale.

\section{Variations on LayerScale init}
\label{sec:variant_layerscale}

For the sake of simplicity and to avoid overfitting per model, we have chosen to do a constant initialization with small values depending on the model depth. 
In order to give additional insight on the importance of this initialization we compare in Table~\ref{tab:init_comp_app} other possible choices. 

\paragraph{LayerScale with 0 init.}
We initialize all coefficients of LayerScale to 0. This resembles Rezero, but in this case we have distinct learnable parameters for each channel. We make two observations. First, this choice, which also starts with residual branches that output 0 the beginning of the training, gives a clear boost compared to the block-wise scaling done by our adapted ReZero. This confirms the advantage of introducing a learnable parameter per channel and not only per residual layer.  
Second, LayerScale is better: it is best to initialize to a small $\varepsilon$ different from zero.

\paragraph{Random init.} 
We have tested a version in which we try a different initial weight per channel, but with the same average contribution of each residual block as in LayerScale. For this purpose 
we initialize the channel-scaling values with the Uniform law ($\mathcal{U}[0,2\varepsilon]$). 
This simple choice choice ensures that the expectation of the scaling factor is equal to the value of the classical initialization of LayerScale. This choice is overall comparable to the initialization to 0 of the diagonal, and inferior to LayerScale. 

\newcommand {\compval} [1] {{\small [{\it #1}]}}
\begin{table}
    \caption{Performance when increasing the depth. We compare different strategies and report the top-1 accuracy (\%) on ImageNet-1k for the DeiT training  (Baseline) with and without adapting the stochastic depth rate $d_r$ (uniform drop-rate), and a modified version of Rezero with LayerNorm and warmup. We compare different initialisation of the diagonal matrix for LayerScale. We also report results with 0 initialization, Uniform initialisation and small constant initialisation. Except for the baseline $d_r=0.1$, we have adapted the stochastic depth rate $d_r$.  
    \label{tab:init_comp_app}}
    \centering
    \scalebox{0.9}{
    \begin{tabular}{c|cc|c|cll}
    \toprule
         \multirow{2}{3em}{\ depth} & baseline & baseline & ReZero & \multicolumn{3}{c}{LayerScale [$\varepsilon$]}  \\
          &  $d_r=0.1$ & [$d_r$] &  $\alpha=0$  &  $\lambda_i=0$ &  $\lambda_i=\mathcal{U}[0,2\varepsilon]$  & $\lambda_i=\varepsilon$\\         
         \midrule
         \pzo12    & 79.9\pzo      & 79.9\  \compval{0.05} & 78.3 & 79.7 & \pzo 80.2 \compval{0.1}     & 80.5 \compval{0.1}     \\ 
         \pzo18    & 80.1\pzo      & 80.7\  \compval{0.10} & 80.1 & 81.5 & \pzo 80.8 \compval{0.1}     & 81.7 \compval{0.1}     \\ 
         \pzo24    & 78.9$\dagger$ & 81.0\  \compval{0.20} & 80.8 & 82.1 & \pzo 82.1 \compval{10$^{-5}$} & 82.4 \compval{10$^{-5}$} \\ 
         \pzo36    & 78.9$\dagger$ & 81.9\  \compval{0.25} & 81.6 & 82.7 & \pzo 82.6 \compval{10$^{-6}$} & 82.9 \compval{10$^{-6}$} \\ 
          \bottomrule
    \end{tabular}}
\end{table}

\paragraph{Re-training.} 

LayerScale makes it possible to get increased performance by training deeper models. At the end of training we obtain a specific set of scaling factors for each layer.   
Inspired by the lottery ticket hypothesis~\cite{frankle2018lottery}, one question that arises is whether what matters is to have the right scaling factors, or to include these learnable weights in the optimization procedure. 
In other terms, what happens if we re-train the network with the scaling factors obtained by a previous training?

In this experiment below, we try to empirically answer that question. We compare the performance (top-1 validation accuracy, \%) on ImageNet-1k with DeiT-S architectures of differents depths. Everything being identical otherwise, in the first experiment we use LayerScale, i.e. we have learnable weights initialized at a small value $\varepsilon$. In the control experiment we use fixed scaling factors initialised at values obtained by the LayerScale training. 

\begin{center}
   \centering \scalebox{0.9}{
    \begin{tabular}{lcccc}
    \toprule
    ~\hspace{3cm} Depth $\rightarrow$              
                                  &  12  &  18  &   24 &  36  \\
    \midrule
    LayerScale                    & 80.5 & 81.7 & 82.4 & 82.9 \\
    Re-trained with fixed weights & 80.6 & 81.5 & 81.2 & 81.6 \\
    \bottomrule
    \end{tabular}
    }
\end{center}

We can see that the control training with fixed weights also converges, but it is only slightly better than the baseline with adjusted stochastic depth drop-rate $d_r$. 
Nevertheless, the results are lower than those obtained with the learnable weighting factors. 
This suggests that the evolution of the parameters during training has a beneficial effect on the deepest models.

\section{Design of the class-attention stage} 
\label{sec:variant_class_attention}

In this subsection we report some results obtained when considering alternative choices for the class-attention stage. 

\paragraph{Not including class embedding in keys of class-attention.} 
In our approach we chose to insert the class embedding in the class-attention: By defining 
\begin{align}
z = [x_\mathrm{class},x_\mathrm{patches}],    
\end{align}
we include $x_\mathrm{class}$ in the keys and therefore the class-attention includes attention on the class embedding itself in Eqn.~\ref{equ:ca2} 
and Eqn.~\ref{equ:ca3}. 
This is not a requirement as we could simply use a pure cross-attention between the class embedding and the set of frozen patches. 

If we do not include the class token in the keys of the class-attention layers, i.e., if we define $z=x_\mathrm{patches}$, 
we reach 83.31\% (top-1 acc. on ImageNet1k-val) with CaiT-S-36, versus 83.44\% for the choice adopted in our main paper. 
This difference of +0.13\% is likely not significant, therefore either choice is reasonable. In order to be more consistent with the self-attention layer SA, in the sense that each query has its key counterpart, we have kept the class embedding in the keys of the CA layers as stated in our paper.

\paragraph{Remove LayerScale in Class-Attention.} 
If we remove LayerScale in the Class-Attention blocks in the CaiT-S-36 model, we obtain a top-1 accuracy of 83.36\% on ImageNet1k-val, versus 83.44\% with LayerScale. 
The difference of +0.08\% is not significant enough to conclude on a clear advantage. For the sake of consistency we have used LayerScale after all residual blocks of the network. 

\begin{table}[t]
    \caption{CaiT models with and without distillation token.
    All these models are trained with the same setting during 400 epochs. \smallskip \label{tab:dist_token}}    
    \centering
    \scalebox{0.9}{
    \begin{tabular}{c|cc}
    \toprule
                & \multicolumn{2}{c}{Distillation token}\\
         Model  & \ding{55} & \checkmark \\
         \midrule
         XXS-24\alambic& 78.4 & 78.5\\
         M-24\alambic & 84.8 & 84.7\\
    \bottomrule
    \end{tabular}}
\end{table}

\paragraph{Distillation with class-attention} 

In the main paper we report results with the hard distillation proposed by Touvron \etal~\cite{Touvron2020TrainingDI}, which in essence replaces the label by the average of the label and the prediction of the teacher output. This is the choice we adopted in our main paper, since it provides better performance than traditional distillation. 

The DeiT authors also show the advantage of considering an additional ``distillation token''. In their case, employed with the ViT/DeiT architecture, this choice improves the performance compared to hard distillation. Noticeably it accelerates convergence.   
 
In Table~\ref{tab:dist_token} we report the results obtained when inserting a distillation token at the same layer as the class token, i.e., on input of the class-attention stage. 
In our case we do not observe an advantage of this choice over hard distillation when using class-attention layers. Therefore in our paper we have only considered the hard distillation also employed by Touvron \etal~\cite{Touvron2020TrainingDI}.

\end{document}